\documentclass[acmsmall]{acmart}

\usepackage{longtable}

\AtBeginDocument{%
  \providecommand\BibTeX{{%
    \normalfont B\kern-0.5em{\scshape i\kern-0.25em b}\kern-0.8em\TeX}}}

\setcopyright{acmlicensed}
\copyrightyear{2026}
\acmYear{2026}
\acmDOI{XXXXXXX.XXXXXXX}

\usepackage{tikz}
\usepackage[dvipsnames]{xcolor}
\usepackage{tabularx,tabulary}

\newcommand{\tikzxmark}{%
\tikz[scale=0.23] {
    \draw[line width=0.7,line cap=round] (0,0) to [bend left=6] (1,1);
    \draw[line width=0.7,line cap=round] (0.2,0.95) to [bend right=3] (0.8,0.05);
}}

\newcommand\BAD{\textcolor{black}{\tikzxmark}}
\newcommand\OKK{\textcolor{ForestGreen}{\checkmark}}

\begin{document}

\vspace*{-8em} 
\begin{center}
\color{red}
\small Published in ACM Transactions on Intelligent Systems and Technology

\url{https://doi.org/10.1145/3831675}
\end{center}

\vspace{1.5em}

\title{Real-time Spatial Retrieval Augmented Generation for Urban Environments}


\author{David Nazareno Campo}
\email{david.campo@alumnos.upm.es}
\affiliation{
  \institution{ETSI de Telecomunicación, Universidad Polit\'ecnica de Madrid}
  \streetaddress{Avda Complutense 30}
  \city{Madrid}
  \state{Madrid}
  \country{Spain}
  \postcode{28040}
}

\author{Javier Conde}
\email{javier.conde.diaz@upm.es}
\author{\'Alvaro Alonso }
\email{alvaro.alonso@upm.es}
\author{Gabriel Huecas}
\email{gabriel.huecas@upm.es}
\author{Joaqu\'in Salvach\'ua}
\email{joaquin.salvachua@upm.es}
\author{Pedro Reviriego}
\email{pedro.reviriego@upm.es}

\affiliation{
  \institution{Information Processing and Telecommunications Center (IPTC), ETSI de Telecomunicación, Universidad Polit\'ecnica de Madrid}
  \streetaddress{Avda Complutense 30}
  \city{Madrid}
  \state{Madrid}
  \country{Spain}
  \postcode{28040}
}


\begin{abstract}

The proliferation of Generative Artificial In\textcolor{black}{t}elligence (AI), especially Large Language Models, presents transformative opportunities for urban applications through Urban Foundation Models. However, base models face limitations, as they only contain the knowledge available at the time of training, and updating them is both time-consuming and costly. Retrieval Augmented Generation (RAG) has emerged in the literature as the preferred approach for injecting contextual information into Foundation Models. It prevails over techniques such as fine-tuning, which are less effective in dynamic, real-time scenarios like those found in urban environments. However, traditional RAG architectures, based on semantic databases, knowledge graphs, structured data, or AI-powered web searches, do not fully meet the demands of urban contexts. Urban environments are complex systems characterized by large volumes of interconnected data, frequent updates, real-time processing requirements, security needs, and strong links to the physical world. This work proposes a real-time spatial RAG architecture that defines the necessary components for the effective integration of generative AI into cities, leveraging temporal and spatial filtering capabilities through linked data. The proposed architecture is implemented using FIWARE, an ecosystem of software components to develop smart city solutions and digital twins. The design and implementation are demonstrated through the use case\textcolor{black}{s} of a tourism assistant \textcolor{black}{and for the management of traffic lights and luminaires} in the city of Madrid. The use case\textcolor{black}{s} serve to validate the correct integration of Foundation Models through the proposed RAG architecture. \textcolor{black}{They} also enable the analysis of current model limitations, such as their inability to handle large volumes of information, even when it fits within their context window, and the high latency of Large Language Models caused by transformer-based architectures, which generate output token by token.

\end{abstract}




\maketitle

\section{Introduction}
\label{sec:Introduction}

The rise of Generative Artificial Intelligence (AI) in general, and foundation models specifically, has enabled the proliferation of new tools in which Large Language Models (LLMs) become just one component within the software architecture. LLMs are neural networks composed of billions of parameters that have been trained on massive datasets with the task of predicting the next token. The transformer-based architecture, along with Big Data capabilities, has significantly increased LLM performance in recent years~\cite{naveed2024comprehensiveoverviewlargelanguage}.


A limitation of foundation models is that their knowledge is restricted to the dataset on which they were trained and the specific task for which they were designed. Additionally, training a model is extremely costly, both computationally and economically, making it unfeasible to train a new model from scratch just to expand its knowledge base. Alternatives like fine-tuning allow us to modify the behavior of the LLMs. Fine-tuning is less expensive than training a model from scratch, and it works well to adapt the behavior of the model by providing a lot of examples of the desired task, but it is less effective in injecting new knowledge~\cite{ovadia-etal-2024-fine}. In addition, it would be necessary to retrain the LLM every time new information needs to be added, making it incompatible with dynamic systems such as cities. Another alternative is Retrieval Augmented Generation (RAG)~\cite{10.5555/3495724.3496517}. RAG is a technique applied to LLMs to provide and restrict the LLM to new information not contained in the training dataset~\cite{ovadia-etal-2024-fine}. RAG is one of the most widely used techniques for integrating LLMs in specific use cases where contextual information, unknown to the base model, is required~\cite{10.1145/3637528.3671470}. Unlike fine-tuning, which modifies the model's parameters through a light retraining, RAG allows external information to be injected without altering the base model. This makes RAG more suitable for dynamic systems with frequent information updates.

One of the environments where base foundation models are not sufficient is cities. Urban environments are characterized by being complex ecosystems with multiple data sources that are updated in real time, which were not in the training dataset of the model and with many interconnected systems. The digitalization of cities through the development of smart cities has been one of the main goals of governments worldwide in recent years~\cite{9086495}, with efforts focused on the creation of urban digital twins (DTs)~\cite{WEIL2023104862}. These urban DTs collect real-time data from the physical city through Internet of Things (IoT) sensors, process them within the virtual entity, and, based on the results, modify the state of the city through IoT actuators. Traditionally, DTs have processed information mainly through physical models or machine learning models aimed at predicting the future state of city elements such as traffic congestion~\cite{auto-2019-0039}. However, these systems have been limited to handling structured data, excluding natural language. The proliferation of Generative AI paves the way for the integration of components into urban environments that are capable of processing natural language. This enables the implementation of more advanced and complex actions designed to improve the lives of citizens, moving toward what is known as Urban General Intelligence~\cite{zhang2025urbangeneralintelligencereview}.

Cities are complex systems that require the integration of multiple data sources, the ability to process large volumes of data in a distributed manner, ensuring information security, interconnecting the physical world through IoT devices, modeling of physical information, processing data in real-time, and the executing complex models~\cite{9086495, exsy.12753, JAVED2022103794}. The need to adapt foundation models to urban environments has led to the emergence of Urban Foundation Models (UFMs), i.e., foundation models that are pre-trained on urban data to be applied in urban applications~\cite{10.1145/3637528.3671453}. These models are designed with capabilities to process natural language, time series, and multimodal information, as well as to perform vision tasks, management, and prediction of human and vehicle trajectories.

The adoption of UFMs requires RAG capabilities in order to process real-time information from cities, with the need of designing new architectures to support all the previously mentioned requirements (connection to the physical world, scalability, security, and real-time processing). In this work, we explore current RAG system solutions for urban environments, propose an architecture that meets the specific requirements of urban environments, and extend a pre-existing digital twin architecture for cities with foundation models. Our proposal is based on interoperable, extensible, open source, and scalable components, enabling the integration of an LLM capable of processing city information. In addition, we include \textcolor{black}{two }use case\textcolor{black}{s} based on the city of Madrid to evaluate the feasibility of the proposed architecture.

The manuscript is structured as follows. In the next section, we discuss existing RAG solutions, their application to urban areas, and we present the FIWARE technology that we will use to implement the urban RAG architecture. In Section 3, we propose the requirements a RAG system must meet to be integrated into cities, as well as a reference architecture to develop urban RAGs. Next, we implement the proposed architecture that uses the component-based technology of FIWARE for cities. In Section 5, we validate our proposal through \textcolor{black}{the mentioned use cases}. Finally, in Section 6, we present the conclusions and limitations of our work.

\section{Preliminaries}

\subsection{Retrieval Augmented Generation}

RAGs are one of the best options for injecting new knowledge into foundation models without the need to retrain it~\cite{ovadia-etal-2024-fine}. The drawback of RAGs is that the new information is not persistent in the model, since it does not modify any of its parameters. Therefore, the maximum amount of information that can be injected is limited by the context window of the model~\cite{NEURIPS2024_1403ab1a}. Additionally, it is necessary to provide the information in each iteration with the model since each prediction starts from the base model. This apparent disadvantage can actually be an advantage for real-time systems, where the value of properties changes over time. If the information was persistent in the LLM, it would need to be able to discern which version of the data is the most recent.

RAGs are used for different purposes, such as answering questions, summarizing texts, text analysis or decision making~\cite{ARSLAN20243781}. If an LLM has been trained up to a certain date, it does not know anything that has happened after that date. For example, LLMs cannot correctly respond to what the weather is like in Paris right now. Another use case of RAGs is to narrow down the context information to the UFM and limit the model's responses to the desired data. For example, with RAG techniques, you can instruct the model on the specifications of a standard or even tell it to respond that the capital of Italy is Naples. Research has shown how RAG helps reduce possible hallucinations by grounding the model response in retrieved facts, as long as the provided information is accurate~\cite{10569238}.

The processing of a naive RAG is divided into different phases~\cite{zhao2024retrievalaugmentedgenerationaigeneratedcontent}. First, there is the indexing phase, where the data are organized into chunks and indexed in an external data source, typically a vector database. This phase ensures that data are stored in small pieces of data, which allows fast search and retrieval based on the user’s query. It is important to design the chunking strategy controlling the way fragments are divided and the size of each chunk~\cite{sarthi2024raptor}. Once the chunks are generated, an embedding is created for each one. At this stage, it is important to choose an embedding model that fits the RAG, observing that higher-dimensional embeddings provide greater search granularity but also slow down retrieval. When a user submits a prompt, it is compared with the stored embeddings. As a first step, the user's prompt can be preprocessed and transformed to improve the search. Next, in the retrieval phase, relevant data is obtained from the indexed source based on semantic similarity metrics. The system performs a search query against the indexed database to retrieve content that matches the user's prompt or context. This phase is crucial to narrow down the most relevant information. There are different retrieval techniques divided into two types of search. In flat searches, all system embeddings are compared. Flat methods offer higher accuracy by exploring all vectors, but are not scalable solutions~\cite{abs-2407-01219}. On the other hand, Approximate Nearest Neighbors (ANN) methods, such as Locality Sensitive Hashing (LSH), Hierarchical Navigable Small World (HNSW), and Inverted File Index (IVF), speed up retrieval by avoiding the need to compare all possible embeddings. Once the data has been received, a reordering of the top-k most relevant documents can be carried out using reranking techniques~\cite{nogueira2020passagererankingbert}. Different data sources can also be merged and the retrieval information can be improved by including neighboring chunks or summaries. 

After the retrieval it starts the content generation phase where the user's prompt and the content extracted during the retrieval and augmentation phase are passed to the LLM. The LLM is then instructed to respond to the user's prompt while limiting itself to the retrieved content. The model generates a response that integrates the information retrieved from the database. Finally, the response generated by the LLM can be refined or corrected. Post-processing can involve tasks such as improving readability, ensuring factual accuracy, or adjusting the tone of the response. The implementation of a RAG is agnostic to the LLM, as the retrieval phase occurs prior to the interaction with the LLM, and the interaction with the LLM is based on a modified prompt that includes the retrieved information and specific instructions for the LLM. For that reason the quality of an RAG depends on both the retrieval architecture but also on the LLM used to generate the response.

Different techniques have also been studied to improve RAGs, taking advantage of advances in embedding models or exploring new paradigms such as token-level embedding, which generates an embedding vector for each token instead of each chunk~\cite{10.1145/3397271.3401075}. However, this type of RAG is not effective with multihop questions~\cite{tang2024multihoprag} in which the solution requires exploring relationships between entities, something that is frequent in urban environments~\cite{10.1145/3637528.3671453}.

To solve this, the researchers propose to model the information as a knowledge graph and take advantage of graph structures to improve the retrieval of information, locate other relevant documents, and generate follow-up questions~\cite{10387715}. Solutions such as GraphRAG~\cite{edge2025localglobalgraphrag} use natural language processing to transform input documents into a graph that contains entities, relationships, and covariates (claims about the entities). To respond to the questions, the information from the embeddings of the chunks, entities, covariates, and relationships is combined with embeddings that represent the structure of the graph itself through graph vector representation algorithms such as Node2Vec~\cite{2939672.2939754}. Graph-based RAGs can apply the same techniques as naive RAGs in the indexing, retrieval, augmentation, and generation phases. These RAGs perform very well for questions involving specific entities (e.g., ``Is Eiffel Tower closed to traffic?''). However, they do not perform well for more general questions (e.g., ``How many streets are currently closed to traffic?''). For global questions, researchers propose generating different levels of summarization that allow global and parallel searches by applying the map-reduce pattern~\cite{edge2025localglobalgraphrag}. RAGs based on knowledge graphs are better suited to urban environments because they are capable of processing relationships. However, real-time information updates can limit their performance, especially with the appearance of new elements that modify the graph structure.

Within RAGs, retrieval is not limited to similarity searches in semantic databases. Solutions that operate on structured data can also be used~\cite{zhao2024retrievalaugmentedgenerationrag}. In this case, an LLM transforms the user prompt in natural language into a query in a specific syntax, for example, some models have been used to generate SQL queries~\cite{10705186}. In this case, the complexity lies not in the search phase, but in the LLM's capabilities to generate SQL queries or extract search parameters from the user's prompt~\cite{10.5555/3666122.3667957}. A key component in these types of solution is the connection with external systems. Proposals such as the Model Context Protocol\footnote{Model Context Protocol: {https://modelcontextprotocol.io/}} facilitate the integration. This approach can be combined with semantic searches, resulting in hybrid systems based on structured and unstructured data sources.

Another popular RAG system is the AI Web searcher or deep research where the retrieval of information comes from Internet searches~\cite{10654534}. This type of RAG allows us to answer dynamic questions such as the current weather in Paris; however, they are limited to public data on the Internet.

In addition to the core processes of retrieval and generation, several factors are crucial in optimizing the performance of RAG systems. Effective prompt engineering is essential to guide the LLM in generating accurate and contextually relevant responses~\cite{10.1007/978-981-99-7962-2-30}. Balancing retrieval accuracy with computational efficiency is another critical consideration, as more complex retrieval strategies can increase latency. While RAG systems provide a flexible and dynamic way to augment the knowledge of LLMs without retraining, they also require careful management of the knowledge base to ensure the reliability and consistency of retrieved information. By addressing these considerations, RAG systems can be effectively utilized to provide real-time, contextually enriched responses in various applications, including urban environments powered by platforms like FIWARE.

Several interesting works have been done with LLM and RAG in the urban domain. Open-TI introduces a system leveraging augmented language models for open traffic intelligence~\cite{DaLiouChen2024OpenTI}. The authors demonstrate the application of advanced language models, potentially using RAG principles, in analyzing traffic data and providing insights for urban transportation management. 
In \cite{smartcities7060121}, authors introduce a digital twin framework for smart-grid energy infrastructure that leverages a RAG pipeline, combining machine learning, a knowledge graph, and an LLM-based conversational assistant; to provide enhanced decision support for asset management and predictive maintenance. Other works \cite{fu2023humanaicollaborativeurbanscience} explore how pre-trained LLMs can facilitate collaborative research between humans and AI in urban science. This work investigates the potential for LLMs to enhance analytical capabilities and address complex urban problems through synergistic interaction between human expertise and AI's processing power.
The study in \cite{ji2023evaluatingeffectivenesslargelanguage} is the first to systematically evaluate foundation LLMs, including GPT-2 and BERT, for encoding geometries in Well-Known Text (WKT) format and preserving spatial relations, demonstrating that while their embeddings can distinguish geometry types and capture spatial relations with up to 73\% accuracy, they still struggle with numeric value estimation and retrieval of spatially related objects, thereby highlighting the urgent need to integrate geospatial domain knowledge to advance GeoAI applications. The work in \cite{10.1145/3631937} is the first to leverage pre-trained models to enhance the candidate-generation phase of Point-of-Interest search in urban environments, substantially improving both retrieval efficiency and the contextual relevance of user recommendations.

Table \ref{tab:rag-summarie} summarizes the characteristics that a RAG must meet for urban environments, as well as the most widely used types of RAGs, indicating whether they meet the requirements. The last column shows real-time spatial RAG, our proposed solution for integrating urban foundation models into urban environments.

\begin{table}[]

\centering
\tiny
\caption{Classification of the different types of RAG based on the requirements of urban environments.}
\vspace{-10pt}
\begin{tabular}{c|ccccc}
\hline
RAG Characterstic & Naive RAG & Graph RAG & \begin{tabular}[c]{@{}c@{}}AI Web \\ Searching\end{tabular} & \begin{tabular}[c]{@{}c@{}}Structured \\ Data RAG\end{tabular} & \textbf{\begin{tabular}[c]{@{}c@{}}Real-time \\ Spatial RAG\end{tabular}} \\ \hline
Contextual data & \OKK & \OKK & \OKK & \OKK & \textbf{\OKK} \\
Private data & \OKK & \OKK & \BAD & \OKK & \textbf{\OKK} \\
Real time data & \BAD & \BAD & \OKK & \OKK & \textbf{\OKK} \\
Spatial filtering & \BAD & \BAD & \BAD & \BAD & \textbf{\OKK} \\
Relationships between entities & \BAD & \OKK & \BAD & \BAD & \textbf{\OKK} \\ \hline
\end{tabular}
\label{tab:rag-summarie}
\vspace{-20pt}
\end{table}

\subsection{FIWARE}

FIWARE is an open source framework that can be assembled together with other third-party components to facilitate the development of smart solutions faster and more efficiently. This includes smart solutions in various domains, such as urban environments. According to the latest information, the platform is in more than 400 cities in at least 35 countries~\cite{fiware2023fiware4cities} and counts more than 630 members in 64 countries, including large corporate, medium and small companies, as well as an ecosystem of innovation centers which are typically run by innovation hubs, RTOs, and universities\footnote{FIWARE Members: {https://www.fiware.org/community/members/organizations-directory/}}.

FIWARE components have served as a foundation for researchers to develop solutions and define architectures in fields such as agriculture~\cite{iotbds23}, industry~\cite{carvajal2024enhancing}, dataspaces~\cite{10.1145/3685651.3686661}, robotics~\cite{agriculture13051005}, or smart cities~\cite{doi:10.1049/PBBE005E_ch10}. The main features of FIWARE components are their interoperability, real-time processing capabilities, integration with external systems and the physical world, security, and scalability, all essential characteristics for smart domains like those mentioned above.

The central element of any FIWARE architecture is the context information, which consists of a set of entities with properties and relationships between them that allows modeling any use case. The core of the FIWARE platform lies in providing tools to manage this context information and can be divided into three fundamental components: 1) the Next Generation Service Interface for Linked Data (NGSI-LD) Standard~\cite{Context_Information_Management} that defines the format of the entities as linked data and provides an API specification to manage them effectively. 2) A suite of components and APIs that enable the integration and management of data from diverse sources through the NGSI-LD standard, facilitating the development of intelligent applications. 3) A collection of ontologies, named Smart Data Models\footnote{Smart Data Models: {https://github.com/smart-data-models}}, that allow model entities and ensure interoperability of data between systems. 

FIWARE components are divided into four layers: a) components responsible for context information management, such as the OrionLD context broker\footnote{OrionLD context broker: {https://github.com/FIWARE/context.Orion-LD}}. This component implements the NGSI-LD standard and provides the ability to access and modify NGSI-LD entities, also allowing asynchronous access to information through a publish-subscribe pattern. b) Components that act as gateways to physical systems or external systems, such as IoT Agents\footnote{IoT Agent: {https://github.com/FIWARE/catalogue/tree/master/iot-agents}}, which enable the connection to IoT sensors and actuators, or Draco\footnote{Draco: {https://github.com/ging/fiware-draco}}, which specializes in connecting to other data sources such as APIs or databases and acts as Extract Transform Load (ETL) system from any protocol and format to NGSI-LD. c) Processing systems, such as Cosmos\footnote{Cosmos: {https://github.com/ging/fiware-cosmos-orion-spark-connector}}, for big data processing and machine learning. d) Auxiliary systems, such as Keyrock\footnote{Keyrock: {https://fiware-idm.readthedocs.io/}} for security. 


Numerous studies have validated FIWARE components for the development of applications in urban environments demonstrating that they meet requirements such as real-time notifications, interoperability between systems, scalability, information security, integration with external systems, and IoT devices~\cite{doi:10.1049/PBBE005E_ch10, CONDE2022101723, ARAUJO2019250, loss2023using, 9346030, 9963753, 10.1007/978-3-031-60796-7-12}. However, none of these works have explored the inclusion of foundation models to enhance and provide new capabilities to these systems.

By combining FIWARE's data management capabilities with the language understanding and generation abilities of LLMs, we can develop intelligent agents, personalized recommendation systems, and content generation tools for urban environments. Moreover, if this can be achieved in real-time and tailored to user preferences, it opens up a new kind of experience for the end user. Our research aims to explore the potential of integrating FIWARE with UFMs for applications in the urban sector. We will extend the FIWARE architecture for digital twins \cite{9346030}, analyze the data flow required for such integration, and evaluate our proposal through a practical use case.

\section{Real-time Spatial RAG for urban environments}

The application of generative AI in urban environments is an idea already explored in the literature~\cite{10.1145/3637528.3671453, DaLiouChen2024OpenTI, smartcities7060121, fu2023humanaicollaborativeurbanscience, ji2023evaluatingeffectivenesslargelanguage, 10.1145/3631937}, which allows adding new functionalities to the city management through queries using natural language, information processing, etc. However, one of the requirements of urban environments is the ability to process data in real-time (or with soft real-time requirements). The concept of real-time in smart cities is broad and encompasses a wide range of scenarios. In systems such as autonomous driving, we deal with hard real-time requirements, where data updates and processing must occur within milliseconds and any delay may cause irreparable damage~\cite{10.1145/3296957.3173191}. In contrast, other systems are less demanding, allowing soft real-time scenarios where information updates can occur in a few seconds~\cite{7474180}. Current architectures based on transformers, which generate information token by token, are not capable of meeting hard real-time requirements so the use of LLMs in urban environments is limited to soft real-time applications~\cite{MLSYS2024_42a452cb}. Additionally, in urban environments, there are a large number of elements, so it is important to have the ability to perform preliminary filtering both in time and space to avoid introducing excessive latency in the retrieval and generation phases~\cite{10.1145/3637528.3671453}. Urban environments are characterized by the existence of multiple relationships between their elements. For example, a malfunction of a traffic light can affect the operation of other nearby traffic lights. Therefore, the RAG system must have navigation and self-discovery capabilities among urban entities. One way to achieve this is by representing information as a knowledge graph, which allows augmenting the LLM context by including new entities~\cite{edge2025localglobalgraphrag}.

In this work, we propose a real-time spatial RAG architecture that satisfies the requirements for urban areas. The spatial dimension refers to the use of the geospatial position as a search criterion. This approach leverages research on clustering techniques, which can keep the number of queries and response sizes controlled. The temporal dimension refers to querying data on databases capable of recording information in real-time. The relationship dimension refers to the ability to retrieve related elements, which can be addressed by modeling the information as a graph. In addition to these three basic features, the RAG architecture must satisfy the other requirements of urban environments:

\begin{itemize}
    \item Connection to the real world. IoT devices enable the connection between the virtual world and the real world. IoT sensors keep the virtual world up to date, while IoT actuators allow actions to be performed in the real world.
    
    \item Interoperability. IoT devices are not the only source of information in urban environments. Other systems can act as data sources, such as databases or APIs (Application Programming Interfaces). The different data sources must follow a common data format to facilitate interoperability. Ideally, these data formats should follow a standard to extend interoperability to external systems.
        
    \item Generation of historical records. Cities generate large amounts of data, so there must be mechanisms for storing historical information for future analysis or to train Machine Learning models. 
    
    \item Scalability. The architectures must be scalable to be able to process hundreds of thousands of data points in real-time.

    \item Security. Urban environments contain sensitive information, making it necessary to protect it through authentication and authorization mechanisms.
\end{itemize}

\begin{figure}
    \centering
    \includegraphics[width=0.4\linewidth]{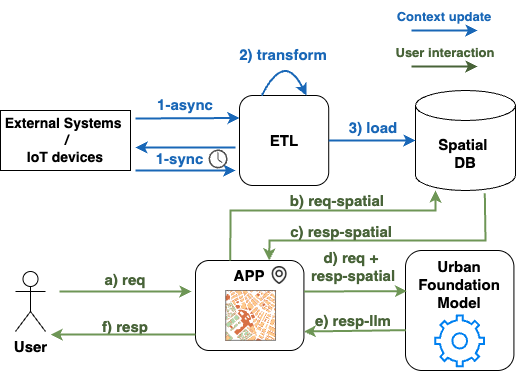}
    \vspace{-15pt}
    \caption{Simplified real-time spatial RAG architecture}
    \label{fig:architecture-general}
    \vspace{-10pt}
\end{figure}

Figure \ref{fig:architecture-general} represents the simplified architecture of a real-time spatial RAG. The information flow starts when an IoT device provides a measurement or an external data source updates the context information (update phase). Data extraction can follow an asynchronous communication flow (1-async), where the data are stored in the database whenever a change occurs at the source. This would allow the system to operate in real-time, which can be implemented through permanent sockets or pattern/subscription architectures. In the case of synchronous communication (1-sync), periodic queries are performed to update the information. In this second approach, the information will not be recorded in real-time. Then a system with ETL capabilities transforms the information into the corresponding data model and creates all the relationships with other entities (2). Finally, in the load phase (3), the linked data are stored in a spatial database that will be accessible by the RAG system. 

In parallel, the RAG system will connect to the data source, and it will be able to process and respond to external prompts with updated information. In this case, the flow begins at (a) when a user makes a request (req) about a region (R). Next, the request is made to the spatial database for the region (R) and any other additional filters specified by the user (b). This interaction with the spatial database can be defined in multiple ways. The number of entities retrieved can be limited and ordered according to defined criteria, linked elements can also be recovered, and clustering techniques can be applied to enhance the scalability of the system. The response from the spatial database (c, resp-spatial) is combined with the initial user prompt (req + resp-spatial) and sent to an UFM (d). The LLM will be configured with a prompt like ``answer the following question: (req) limited to the data (resp-spatial)'' resulting in the UFM response (resp-llm) adjusted to the data extracted from the spatial database (e). Finally, the final response (resp) is composed and sent to the user (f). When a user makes a query again, a new interaction with the RAG will begin (a-f). Since flows (1-3) and (a-f) run in parallel, the system is able to react to changes in real-time. The capabilities of spatial databases allow the response size to be limited to the established region, the temporal filtering allows filtering data by date, and the linked data helps to improve the retrieval by including related entities. 

\textcolor{black}{This work presents the implementation of a reference architecture capable of operating under the real-time and physical-world interaction requirements typical of smart cities, aligned with the definition of Urban Foundation Models proposed by Zhang et al. \cite{zhang2025urbangeneralintelligencereview}. In contrast to the work of Ji and Gao \cite{ji2023evaluatingeffectivenesslargelanguage}, which focuses on evaluating the intrinsic ability of LLMs to encode geometric text (WKT) and spatial relationships in isolation, the FIWARE-based approach shifts the complexity of spatio-temporal filtering to a standardized Context Broker (NGSI-LD) prior to the generation phase. This approach introduces a key difference with respect to traditional RAG architectures, such as Naive RAG \cite{zhao2024retrievalaugmentedgenerationaigeneratedcontent} or Graph RAG \cite{edge2025localglobalgraphrag}: the incorporation of a spatial RAG mechanism that filters entities based on their geographic location. By doing so, it reduces the amount of context provided to the model and addresses common issues related to scalability, latency, and hallucinations when LLMs process large volumes of dynamic urban data. Other studies, such as those proposed by Conde et al. \cite{9963753} or Araujo et al. \cite{ARAUJO2019250}, already suggest the use of FIWARE for smart cities, but they have not explored the integration of LLMs or RAGs.}

\textcolor{black}{Recent research has explored various ways to enhance the capabilities of LLMs for urban data management. Feng et al. \cite{feng2025urbanllava} propose fine-tuning of multimodal models to improve the understanding of urban imagery sourced (e.g., maps, satellites). In other work they propose to adjust model weights through urban-specific instruction tuning what enables smaller models to achieve competitive performance in urban management tasks \cite{10.1145/3711896.3736878}. Wang et al. \cite{wang2025urban} propose optimizing models via reinforcement learning using the GRPO algorithm to strengthen reasoning in complex urban scenarios. While these works enhance the model's intelligence, they do not address the technical integration with dynamic city data. However, these approaches are complementary to our architecture, which remains agnostic to the specific LLM utilized.}

\textcolor{black}{Regarding spatial information retrieval, Zhu et al. \cite{zhu2025boundary} investigate filtering through elastic regions instead of fixed grids by generating graphs with spatial tokenizers. This is a promising approach, but it introduces additional latency compared to our architecture based on Linked Data, where the graph is already defined internally. Jin et al. \cite{10.1145/3711896.3737176} train graphs designed to extract urban behavior patterns and resolve complex queries, however, these models show limitations in adapting to constant, real-time changes in the city’s state.
}

\textcolor{black}{The work of Yu et al. \cite{yu2025spatial} proposes a framework that balances spatial and semantic relevance via a hybrid retriever for reasoning queries. Our RAG-FIWARE architecture differs fundamentally by focusing on real-time operability and direct interaction with the physical world through IoT devices. Unlike the approach in \cite{yu2025spatial}, our proposal leverages an asynchronous subscription-based architecture under the NGSI-LD standard, allowing it to react to dynamic urban changes (e.g., traffic states or occupancy levels) without the overhead of constant data re-indexing. By delegating spatial and temporal filtering to the FIWARE Context Broker prior to generation, we significantly reduce the LLM’s context load and ensure interoperability via Smart Data Models. Ultimately, by relying on standardized technologies already deployed in urban ecosystems, our architecture ensures fluid integration and immediate operational readiness for large-scale, real-world deployments.}

\textcolor{black}{Our} real-time spatial RAG architecture meets all the requirements for the effective integration of UFM in urban environments. Table \ref{tab:rag-summarie} presents a comparison with other popular RAG systems in the literature. 


\section{Implementation of Real-Time Spatial RAG with FIWARE}


Recently, the integration of LLMs with different tools like LlamaIndex\footnote{LLamaIndex: https://www.llamaindex.ai/}, Langchain\footnote{LangChain: https://www.langchain.com/}, GraphRag\footnote{GraphRag: https://microsoft.github.io/graphrag/} has gained significant attention for the development of RAGs. However these tools are focused on RAGs for retrieving information from documents or graphs and do not meet all the requirements for urban environments. In this work, we propose an architecture based on FIWARE to bridge this gap.

\begin{figure}
    \centering
    \includegraphics[width=0.9\linewidth]{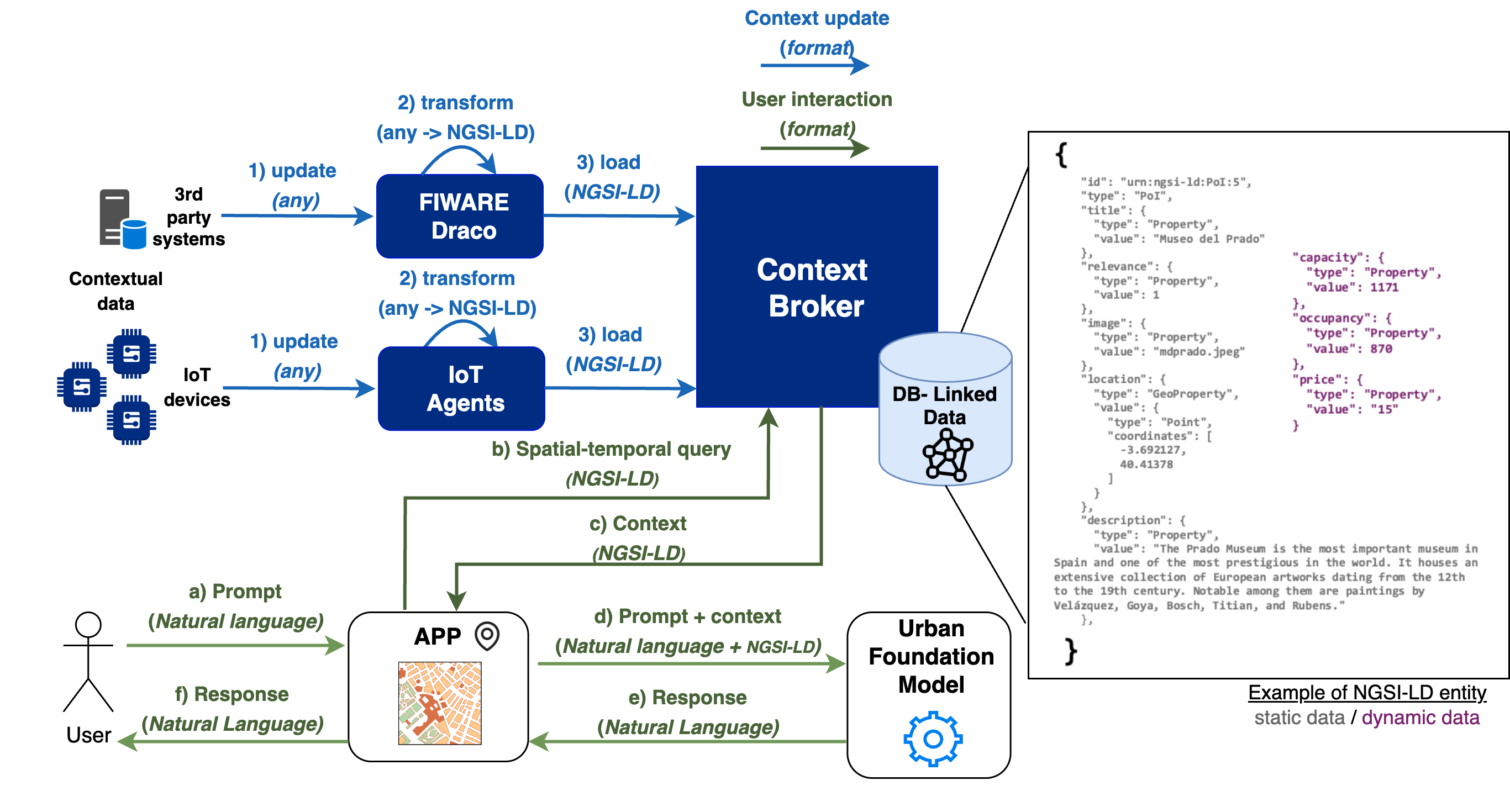}
    \vspace{-15pt}
    \caption{General proposed architecture implemented with FIWARE}
    \label{fig:architecture}
    \vspace{-20pt}
\end{figure}

Figure \ref{fig:architecture} illustrates a general architecture that integrates FIWARE technology with UFMs to provide a real-time spatial RAG. This architecture aims to deliver contextually enriched real-time information from entities in a city to be processed by UFMs. The centerpiece of \textcolor{black}{the} architecture lies the context broker, which serves as the component responsible for managing and storing contextual information. The context broker utilizes a database system for persistent storage and can link to other data repositories, including binary data such as images. Data can be populated into the system through various methods. Users can inject raw entities directly into the context broker using NGSI-LD POST queries, such as storing Points of Interest (PoI) entities of a city. In addition, relevant contextual information can be incorporated from third-party systems. FIWARE Draco normalizes and transforms the data from these systems into NGSI-LD format before sending it to the context broker and making them compliant with the ontologies defined by the Smart Data Models. IoT agents play a role in receiving data from various IoT sensors and devices, converting it into NGSI-LD format and updating the information in the context broker. External data sources, such as weather updates or social media feeds, can also be registered to provide real-time information, enhancing the contextual data model.

As illustrated in Figure \ref{fig:architecture}, a user can interact with the system through an application. This application displays contextual information and reflects changes resulting from user interactions or updates from other systems. When a user submits a question about the city, the RAG application retrieves relevant entities from the context broker using the NGSI-LD API, filtering by entity properties, date information, spatial location, ensuring that the response is geographically, temporally, and contextually relevant, and optionally retrieving additional entities that are linked through NGSI-LD relationships. The application then combines the retrieved contextual data with the user's prompt and sends this enriched prompt to an LLM. The LLM processes the prompt, leveraging its knowledge base and the provided context to generate an informative and contextually appropriate response. This response is returned to the application, which presents it to the user in a user-friendly format. Thanks to the subscription system of the context broker, the RAG can be configured to receive real-time updates of modified entities, eliminating the need for periodic requests to check whether the contextual information provided to the LLM has changed, something common in urban environments.

In the initial experiments that we will present later, we observed that LLMs are capable of understanding the NGSI-LD format, which simplifies the post-processing of the retrieved entities. In cases where the LLM is unable to interpret the properties of an NGSI-LD entity, it can be programmed to consult the ontology, as NGSI-LD entities include a ``context'' attribute that links to the Smart Data Model used. This model provides a natural language description of the entity and explains the meaning of each of its properties and relationships.

\textcolor{black}{
\section{Results and use cases}
}
In this section, we present \textcolor{black}{two} implementation example\textcolor{black}{s} in Madrid that will allow us to validate the proposed architecture \textcolor{black}{and compare it with other RAG solutions}. To do so, we will evaluate the system's performance in different scenarios, considering \textcolor{black}{the following} dimensions: response speed, response correctness, \textcolor{black}{precision, and recall}. All code, data sets, and results are available in a public zenodo repository\footnote{https://doi.org/10.5281/zenodo.20889036}.

\subsection{Definition of the use case\textcolor{black}{s}}

The \textcolor{black}{first} use case consists of developing a real-time tourist assistant for the city of Madrid. To this end, 1,088 PoIs have been loaded, including both static information (e.g., name and description) and dynamic information (e.g., visitor affluence and price). This use case allows us to validate the architecture, as it enables the delivery of real-time, contextually enriched information about the city's points of interest, ensuring that tourists receive up-to-date and relevant insights. \textcolor{black}{PoI entities have a unique identifier, a title, a relevance level on a scale from 1 (very famous) to 5 (not very relevant), a geospatial location (WGS-84, point geometry), a price (a value or a range), a description of the place, a maximum capacity and a current occupancy level (both in number of people).} \textcolor{black}{The second use case consists of a management system for traffic lights and luminaires, 650 of each, which include static information such as name, description, and location, as well as dynamic information, such as, status and density of people around them.} \textcolor{black}{Luminaires entities encode the operational state and local human activity context of street lights. Each record includes a unique NGSI-LD identifier, administrative metadata (district and neighborhood), the current status (on, off, broken), a simple crowd-estimate attribute (people count), and precise geographic coordinates (WGS-84, Point geometry)}.\textcolor{black}{TrafficLightSignal entities describe traffic light systems deployed across the city. Each entry includes a unique NGSI-LD identifier, district, an installation date, a textual description of the intersection, the current signal state (red, yellow, green), and a geospatial location (WGS-84, Point geometry).}

To ensure clarity and facilitate understanding, Table \ref{tab:correctness_results_limit10} shows some examples of entities and their corresponding attributes. Certain attributes, such as the description \textcolor{black}{in PoIs}, were intentionally omitted to maintain focus on the most relevant and concise information.

{\tiny
\begin{longtable}[c]{|l|cc|c|c|c|}
\caption{\textcolor{black}{Examples of NGSI-LD entities}}
\vspace{-10pt}
\label{tab:correctness_results_limit10}\\
\hline
\textbf{PoI ID}          & \multicolumn{1}{c|}{\textbf{Title}} & \textbf{Relev.} & \textbf{Price}    & \textbf{Capacity} & \textbf{Occupancy}    \\ \hline
\endfirsthead
\endhead
PoI:23                  & \multicolumn{1}{l|}{Hospital San Carlos} & 1 & 0€     & 1679   & 1067       \\
PoI:170                 & \multicolumn{1}{l|}{Restaurante StreetXO}        & 1 & 60-80€ & 578    & 523        \\ \hline
\textbf{\textcolor{black}{TrafficLight ID}} & \multicolumn{2}{c|}{\textbf{\textcolor{black}{Description}}}                & \textbf{\textcolor{black}{District}} & \textbf{\textcolor{black}{Status}}   & \textbf{\textcolor{black}{Installation}} \\ \hline
\textcolor{black}{TrafficLightSignal:101}  & \multicolumn{2}{l|}{\textcolor{black}{Princesa - Romero Robledo}}       & \textcolor{black}{9}      & \textcolor{black}{red}    & \textcolor{black}{16/06/1962} \\
\textcolor{black}{TrafficLightSignal:1096} & \multicolumn{2}{l|}{\textcolor{black}{Entrevias - Mendez Alvaro}}   & \textcolor{black}{13}     & \textcolor{black}{green}  & \textcolor{black}{17/09/1984} \\ \hline
\textbf{\textcolor{black}{Luminaire ID}}     & \multicolumn{2}{c|}{\textbf{\textcolor{black}{Neighborhood}}}               & \textbf{\textcolor{black}{District}} & \textbf{\textcolor{black}{Status}}   & \textbf{\textcolor{black}{People}}       \\ \hline
\textcolor{black}{Luminaires:149}          & \multicolumn{2}{c|}{\textcolor{black}{4}}                               & \textcolor{black}{16}     & \textcolor{black}{on}     & \textcolor{black}{1}          \\
\textcolor{black}{Luminaires:23000}       & \multicolumn{2}{c|}{\textcolor{black}{4}}                               & \textcolor{black}{13}     & \textcolor{black}{broken} & \textcolor{black}{4}          \\ \hline
\end{longtable}
\vspace{-5pt}
}

We have implemented a similar but simplified architecture as depicted in Figure \ref{fig:architecture}. The system employs the FIWARE OrionLD context broker to manage NGSI-LD data related to POIs, \textcolor{black}{traffic lights and luminaires}. This data management ensures that all information is up-to-date and accurately reflects real-time conditions. An example of a POI NGSI-LD entity can be consulted in Figure \ref{fig:architecture}. \textcolor{black}{Text embeddings were generated using the text-embedding-3-large\footnote{\textcolor{black}{https://platform.openai.com/docs/models/text-embedding-3-large}} model with 1024 dimensions, which was employed for the initial retrieval stage. The retrieved results were then reranked using the ms-marco-MiniLM-L-6-v2\footnote{\textcolor{black}{https://huggingface.co/cross-encoder/ms-marco-MiniLM-L6-v2}} model to improve relevance. We used Qdrant as the vector database} 


The system includes an interactive map that allows users to explore specific areas of the city. The map dynamically updates the bounding box coordinates, which are used to filter relevant \textcolor{black}{entities} from the context broker. This interactive element helps users focus on areas of interest and obtain detailed information about these locations while limiting the number of \textcolor{black}{entities} retrieved. Users submit questions through a text box that functions as a chat interface. These queries can be specific questions about a particular monument or broader inquiries about nearby attractions and general travel tips. The user's question, along with the retrieved data from the context broker are sent to the LLM, which acts as an expert in NGSI-LD. We have tested \textcolor{black}{GPT5.2 (gpt-5.2-2025-12-11)}, GPT-4.1 (gpt-4.1-2025-04-14), \textcolor{black}{GPT-4.1 mini (gpt-4.1-mini-2025-04-14), Llama3.1-8B-instruct, Llama3.1-70B-instruct, Qwen2.5-7B-instruct and Deepseek-V3-685B}, as \textcolor{black}{they are ones of the most powerful models currently available, offer a large context window, and constitute a diverse sample of large and small models, both commercial and open-weighted}. The LLM synthesizes the provided data with its internal knowledge base to generate a comprehensive and informative response. The model temperature has been set to 0 to ensure that the experiment is reproducible. \textcolor{black}{Prompts were defined based on an exploratory process using different techniques. We arrived at the best configuration with a detailed system prompt that instructs the model on the role it should assume, how it should behave, the expected output format, and what to do when it cannot find entities that satisfy the user’s questions. The system prompt can be adapted to each scenario and model, as prompt engineering remains an important phase in the setup of RAGs and the use of LLMs. After the system prompt, the entities are provided to the LLM in NGSI-LD format along with the user’s question. It has also been tested to change the order (system prompt + user question + entities), but no differences have been observed}:
\begin{itemize}
    \item System prompt, specific for each use case:
    \begin{itemize}
        \item \textcolor{black}{Tourism use case}: ``You are a tourist guide in the city from where the data is provided. Also you are expert in NGSI and semantics. You should only answer about the following points of interests that I will provide you in NGSI format. At first, you should only provide the name of the places with not extra detail, unless requested in the prompt message by the user. If you don't know about any place, or you cannot find anything matching the request you should just say that you can't find anything in a expressive and emphatic way related to the asked question. You should provide the information in plain text, with natural language understandable by tourists. Please, also consider only the following points of interest when giving advices. Otherwise, just say that you can't find anything. Answer in plain natural text please.''
        
        \item \textcolor{black}{Traffic lights / Luminaires use case: ``You are expert in NGSI and semantics.  You should only answer about the following entities that I will provide you in NGSI format. At first, you should only provide the name of the entities with not extra detail, unless requested in the prompt message by the user. If you don't know about any place, or you cannot find anything matching the request you should just say that you can't find anything related to the asked question. You should provide the information in plain text, with natural language understandable by a human. Please, also consider only the following points of interest when giving advices. Otherwise, just say that you can't find anything. Answer in plain natural text please, no markdown nor HTML. Simple items including only the title, unless requested by the user. Please, If there are NO entities, DO NOT GIVE ANY HINT, just say you do not know.''}
    \end{itemize}

    \item Entities: \{\{list of entities from the context broker \textcolor{black}{in NGSI-LD format as the ones from Figure \ref{fig:architecture} provided after the spatial filtering, the retrieval and the rerank}\}\}
    \item User prompt: \{\{specific question\}\}.
\end{itemize}

The system continuously updates information based on real-time data changes, such as visitor affluence or price, ensuring that users will receive the most current and relevant information. This dynamic capability is crucial for providing accurate and timely guidance to \textcolor{black}{users}. The subscription system provided by FIWARE enables the RAG to meet the real-time requirements of smart cities. A typical workflow would involve an IoT sensor updating an entity in the context broker (e.g., the occupancy level of a landmark). The context broker would then automatically send a notification to all users subscribed to that entity. In our scenario, the tourist assistant would immediately make the updated information available to the UFM and the UFM would be able to answer the user in natural language. 

Initially, NGSI-LD \textcolor{black}{entities} are loaded into the FIWARE context broker \textcolor{black}{and their respective embeddings to Qdrant}. One limitation of urban environments is the high number of elements, which makes searches in a naive RAG that lacks geographic search capabilities, both challenging and slow. To overcome this, the assistant leverages the geospatial and temporal filtering capabilities of the FIWARE context broker to reduce the number of entities. The number of retrieved entities directly affects the RAG’s performance, both in terms of execution time and response accuracy. In the \textcolor{black}{two use cases}, the filtering is applied based on the geographical area of interest specified by the user. In this study, rectangular polygons were used.

\begin{figure}
    \centering
    \includegraphics[width=0.6\linewidth]{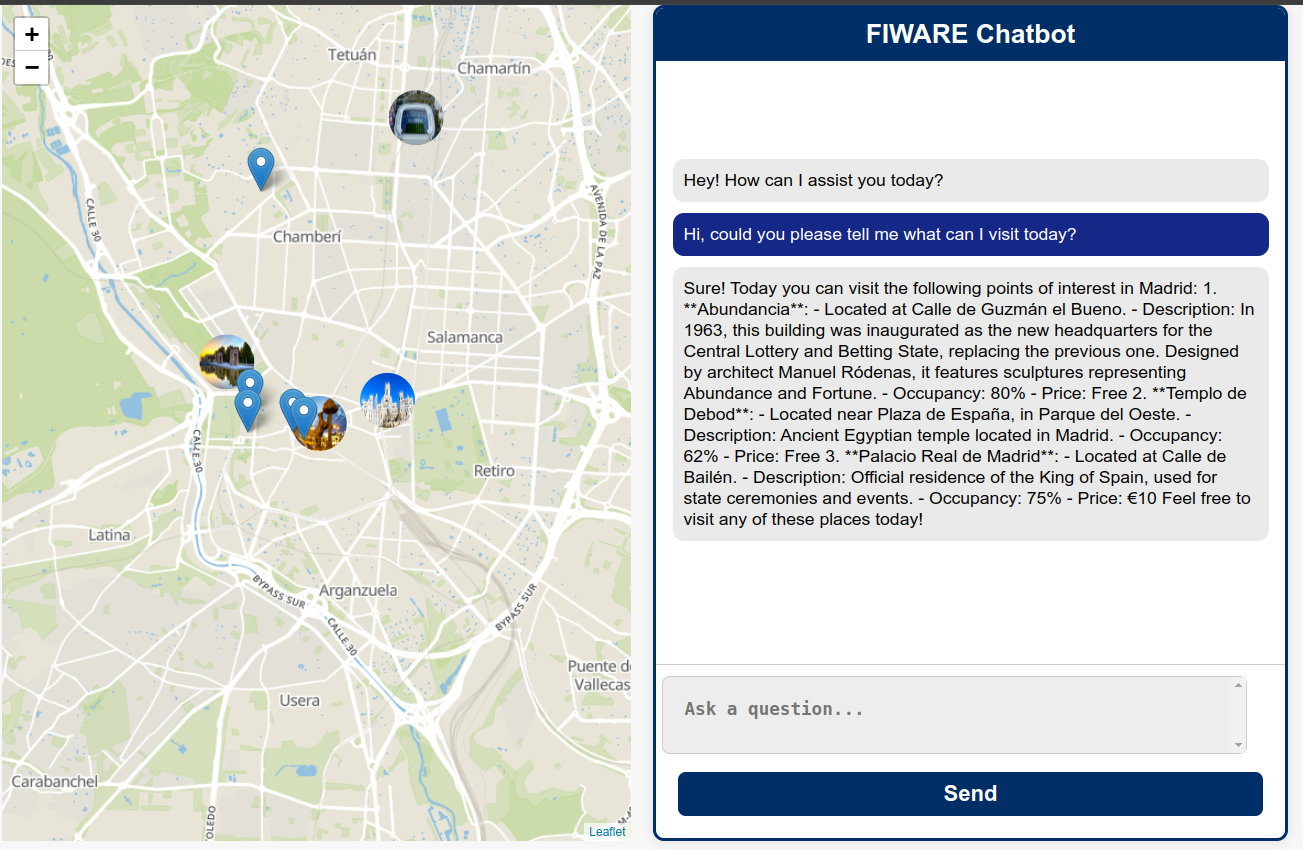}
    \caption{Screenshot depicting a user asking recommendations to the application}
    \vspace{-5pt}
    \label{fig:app}
\end{figure}

A tourist planning to visit a famous monument in Madrid can use the application to find optimal visit times, current price conditions, and nearby events. By zooming into the area of interest on the map, they receive detailed information about monuments, including their history and significance, enriched by the UFM to provide an engaging narrative. This is depicted in Figure \ref{fig:app} where a screenshot of the interface shows a user asking for recommendations. \textcolor{black}{A similar pipeline is followed in the case of an operator who wants to manage the state of the streetlights and traffic lights.}


The process begins with the user interacting with a map on the front end, performing actions such as zooming in or moving. When the user asks a question, the front end sends a geo query to the context broker with the current map coordinates, which returns the current \textcolor{black}{entities} placed in the current zoom and position parameters. The front end then forwards the user's question along with the \textcolor{black}{NGSI-LD entities} to the LLM API for processing. The LLM API returns an answer, which the front end displays to the user. \textcolor{black}{Spatial RAG can be combined with semantic RAG by introducing a retrieval phase restricted to entities within the selected area. In addition, reranking can be applied to further refine the set of entities provided to the LLM.} For the sake of simplicity of the evaluations, we are not using either IoT agents or Draco because we have preloaded all the entities in the context broker. This simplification does not affect the results and the validation of IoT agents and Draco in smart cities has already been addressed in other studies \cite{10.1007/978-3-031-60796-7-12, 9346030}.


\subsection{Results and discussion}

We devised a \textcolor{black}{three}-stage experimental setup in order to evaluate the efficiency and the quality of the responses generated by our LLM-driven RAG architecture. In the \textcolor{black}{first} stage, we conducted a series of tests to assess the ability of the model to provide correct and contextually relevant answers. \textcolor{black}{For this purpose, we tested three regions of increasing size: Small (10 - 20 entities), Medium (100 - 200) and Large (650 - 1300). While these quantities may fit within their context windows, they can still limit the models’ ability to process the information effectively. We selected a representative set of LLMs, including both commercial and open-weight models of different sizes: GPT-5.2, GPT-4.1, GPT-4.1-mini, LLama 3.1 70 B, LLama 3.1 8B, Deepseek V3 685B and Qwen 2.5 7B. The use of commercial models versus open-weight models that can be run locally responds to different operational and contextual needs. Commercial models are generally preferable when maximum performance is required in terms of quality, complex reasoning, and robustness, as well as rapid access to state-of-the-art models. This makes them especially suitable for environments with stable connectivity and where network latency is not critical. In contrast, local models offer clear advantages in scenarios with no connectivity, strict privacy or data sovereignty requirements. They also enable offline operation and predictable latency.} \textcolor{black}{In the second stage, we validated our proposal across different architectures by conducting an evaluation that compares with other common RAG solutions: (1) Semantic RAG; (2) Semantic RAG with reranking; (3) Spatial RAG; (4) Spatial RAG combined with Semantic RAG; and (5) Spatial RAG combined with Semantic RAG and reranking.} \textcolor{black}{Finally, in the third stage we measured} the response times associated with the retrieval and subsequent LLM processing, examining how variations in query parameters, \textcolor{black}{configurations,} and data volume influence latency.

To this end, we crafted a set of \textcolor{black}{24} distinct questions, each representing a different query type \textcolor{black}{derived from the two datasets presented previously (tourism and traffic lights and luminaires)}. To ensure consistency in the results, no modifications were introduced to the bounding box defining the search area, so all queries targeted the same fixed geographical region. Due to the deterministic nature of the context broker, repeated queries with the same parameters yielded an identical set of entities, thus minimizing variability arising from data retrieval.

Tables~\ref{tab:ground_truth_pois} and \textcolor{black}{\ref{tab:table-q-streetlights-luminaires}} collect all the questions about tourism (QT) and \textcolor{black}{traffic lights and luminaires (QL)} and for each of them the number of entities that address under the \textcolor{black}{three} different regions: \textcolor{black}{Small, Medium and Large}. Each row corresponds to one question, and the entries in each column indicate the number of entities retrieved by the context broker that met the query’s criteria. For instance, when the retrieval limit is set to 10, certain queries (e.g., QT1) match all the available entities. Conversely, other queries (e.g., QT5 under limit 10) yield no matching PoIs in that limited subset, making \textcolor{black}{the only meaningful response that it does not know the answer}. 

\begin{table}[ht]
\centering
\tiny
\setlength{\tabcolsep}{6pt}
\renewcommand{\arraystretch}{1.2}
\caption{Questions to evaluate the correctness of the real-time spatial RAG system and the number of PoIs in the \textcolor{black}{tourism} dataset that satisfy the criteria of the corresponding question for the given region limit: Small (10--20 entities), Medium (100--200 entities), Large (650--1300 entities)}
\vspace{-5pt}
\begin{tabularx}{\textwidth}{c|X|ccc}
\hline
 &  & \multicolumn{3}{c}{\textbf{Region size}} \\
\textbf{QT\#} & \textbf{Question} & \textbf{Small} & \textbf{Medium} & \textbf{Large} \\
\hline

QT1 & Could you recommend the most interesting landmarks or places in Madrid to me? & 10 & 100 & 650 \\

QT2 & Please, show me a list of the top 5 most relevant sites in Madrid & 10 & 29 & 29 \\

QT3 & Please, inform me of all the places I can visit in Madrid today that have a cost between 10€ and 20€ & 1 & 5 & 7 \\

QT4 & Please, show me some landmarks that are free of charge & 8 & 90 & 599 \\

QT5 & Please, list some places that are related to sports & 0 & 1 & 1 \\

QT6 & Do you know the Museo del Prado? & 0 & 1 & 1 \\

QT7 & Could you let me know if the Museo del Prado is free of charge to enter? & 0 & 1 & 1 \\

QT8 & Please tell me if the Museo del Prado is currently crowded? & 0 & 1 & 1 \\

QT9 & Could you show me places with an occupancy of less than 50 people and a relevance of 1? & 0 & 3 & 3 \\

QT10 & Could you show me places that have an occupancy of not less than 50 people and a relevance that is not 1? & 0 & 64 & 546 \\

QT11 & Could you show me places with an occupancy of less than 50 people or a relevance of 1? & 10 & 36 & 104 \\

QT12 & Could you show me places that are occupied by not fewer than 50 people or have a relevance not equal to 1? & 10 & 97 & 647 \\

\hline
\end{tabularx}
\vspace{-6pt}

\label{tab:ground_truth_pois}
\end{table}

\begin{table}[t]
\centering
\tiny
\setlength{\tabcolsep}{4pt}
\caption{\textcolor{black}{Questions to evaluate the correctness of the real-time spatial RAG system and the number of entities in the traffic lights and luminaires dataset that satisfy the criteria of the corresponding question for the given limit: Small (10-20 entities), Medium (100-200 entities), Large (650-1300 entities)}}
\vspace{-6pt}
\label{tab:table-q-streetlights-luminaires}
\color{black}
\begin{tabularx}{\textwidth}{c|>{\raggedright\arraybackslash}X|cccc}
\hline
 &  & \multicolumn{4}{c}{\textbf{Region size}} \\
\textbf{QL\#} & \textbf{Question} &
\textbf{Small} & \textbf{Medium} & \textbf{Large} \\ \hline
QL1 & What is the current status of the traffic light at the intersection of TORRELAGUNA and ARTURO BALDASANO? & 1 &  1  &  1 & \\ \cline{2-2}
QL2 & Please show me a list of all traffic lights that are currently `Red`. & 4 & 45 & 287 \\ \cline{2-2}
QL3 & Identify all traffic lights currently in the `Yellow` state. & 3 & 8  & 66 \\ \cline{2-2}
QL4 & List all the intersections involving ESTACION HORTALEZA that have a traffic light installed. & 0 & 3  & 3 \\ \cline{2-2}
QL5 & Please identify all luminaires that are currently marked as `Broken`. & 1 & 7  & 69 \\ \cline{2-2}
QL6 & List all luminaires with no affluence of people. & 3 & 39 & 255 \\ \cline{2-2}
QL7 & Which luminaires have an affluence of less than 5 people? & 10 & 100  & 646 \\ \cline{2-2}
QL8 & Identify entities where the traffic light status is not `Green' and the closest luminaire within 10000 meters either has an affluence value below 5 or has a bulb type equal to `LED'. & 17 & 153 & 999 \\ \cline{2-2}
QL9 & Display entities where the traffic light is `Yellow' OR the luminaire within 100 meters is `Broken' but NOT where affluence is lower than 4 & 13 & 108 & 717 \\ \cline{2-2}
QL10 & List all intersections where the traffic light is neither `Red' nor `Green'. & 3 & 8  & 66 \\ \cline{2-2}
QL11 & Based on the data, create a list of locations that require immediate repair. & 1 & 7  & 69 \\ \cline{2-2}
QL12 & Compare the number of `Red` traffic lights to the number of `Broken' luminaires in the provided area. & 5 & 52 & 356 \\ \hline \cline{2-2}
\end{tabularx}

\end{table}

\subsubsection{Assessing correctness on different contextual information}
\label{ref_results_correctness} 
For the first set of experiments, we maintain the same foundational setup as described in the previous section. Once again, we used the FIWARE OrionLD context broker and \textcolor{black}{different LLMs.}


This round of experimentation sought to evaluate only the correctness and contextual quality of the answers provided by the UFM. These new prompts were designed to test more rigorously the processing capabilities of the model. By adjusting the questions, while maintaining the underlying data source and processing pipeline unchanged, our aim was to isolate the factors that influence both the accuracy of the LLM's output and the reliability of its reasoning, ultimately gaining deeper insight into how the model handles real-time data when correctness is a primary objective.

The questions range from broad, open-ended requests for general recommendations to more intricate inquiries and combined logical conditions. In this batch of experiments, we designed the queries to progressively assess the ability of the LLM to interpret contextual data and apply logical constraints. 

\textcolor{black}{Left panels of Figure \ref{fig:recall_precision} present the qualitative assessment of LLM responses. Results are aggregated over both question sets (QT and QL) and reported as the percentage of questions assigned to each qualitative category per model and region size. \textcolor{black}{For the evaluation we employed a human-in-the-loop methodology conducted by experts on the NGSI-LD standard, the Smart Data Models, the specific datasets used, and familiar with the City of Madrid, guaranteeing that the evaluation covers all critical dimensions, from technical format compliance to geographical accuracy. Evaluators have to clasify each response in one of the following categories}: OK corresponds to correct answers with a hit of True Positive (TP) values of at least 70\% with no hallucinated entities; MED to correct answer but hit of TP between 30\% and 70\% with no hallucinations; BAD to a hit of TP below 30\% with no hallucinations; HALLUCINATES to any answer containing at least one invented entity regardless of amount of TP hit; and ERROR to any runtime, API, or context-related failure. Notably, the hallucination criterion is intentionally strict: a single false positive is sufficient to classify an answer as HALLUCINATES. This methodological choice strongly penalizes verbose or over-complete responses.}
\textcolor{black}{The ERROR category appears exclusively for Llama-3.1-8B-Instruct and Qwen-2.5-7B-Instruct, and only in the Large scenario, due to lack of context window. Consequently, ERROR should be interpreted as a technical limitation rather than a cognitive failure, and it highlights the importance of context window size for large-scale retrieval tasks.}

\textcolor{black}{Across nearly all models, the proportion of OK answers decreases as region size increases. This reflects the growing difficulty of simultaneously achieving high TP hit while avoiding hallucinated entities as the number of relevant targets expands. Larger regions impose a more severe precision–recall tradeoff, increasing the likelihood of false positives and, therefore, hallucination classifications. This trend indicates a general scalability limitation in multi-entity retrieval and reasoning tasks, where maintaining both completeness and correctness becomes progressively harder as the context grows.}
\textcolor{black}{In contrast, the MED category remains relatively stable across region sizes for most models. This suggests that maintaining moderate TP recall without hallucinations is more feasible than achieving high TP recall without introducing false positives. As a result, MED effectively represents a consistent intermediate performance regime when optimal performance degrades, capturing cases where models retrieve a meaningful subset of relevant entities while remaining conservative enough to avoid hallucinations.}


\textcolor{black}{GPT-4.1-mini achieves the highest OK ratios in both the Small and Medium scenarios, with better results than theoretically better models like GPT-4.1 or GPT-5.2 , indicating strong performance in moderate-context retrieval tasks. However, its performance deteriorates sharply in the Large scenario, primarily due to increased hallucinations and reduced TP recall. This suggests that GPT-4.1-mini is optimized for compact contexts but lacks robustness for large-scale retrieval, where the complexity of the task exceeds its effective reasoning or context-handling capacity.}
\textcolor{black}{GPT-5.2 demonstrates the most consistent performance across all region sizes. While it closely follows GPT-4.1-mini in the Small and Medium scenarios, it clearly outperforms all models in the Large scenario. GPT-5.2 exhibits higher BAD rates but lower HALLUCINATES rates, indicating a conservative response strategy that prioritizes hallucination avoidance over TP hit. In practice, the model favors shorter, safer answers, accepting lower TP recall to minimize FP. This precision-oriented strategy proves advantageous at scale, where the risk of hallucinations increases substantially.}
\textcolor{black}{The contrast between GPT-5.2 and GPT-4.1-mini reveals two distinct response strategies. GPT-4.1-mini emphasizes recall and answer completeness in small and medium contexts but is sensitive to scale, whereas GPT-5.2 seems emphasize hallucination control and maintains robustness as context size increases. This tradeoff underscores the importance of aligning model selection with task requirements, particularly when large retrieval contexts are involved.}
\textcolor{black}{Among open-weight models, DeepSeek-V3-685B achieves the strongest qualitative performance, with OK ratios lower than GPT-5.2 but comparable to it. Its strong results are likely attributable to its large parameter count, advanced instruction tuning, and improved long-context handling. Llama-3.1-70B-Instruct ranks second among open models, showing moderate scalability with some hallucination risk. Smaller open-weight models, such as Llama-3.1-8B-Instruct and Qwen-2.5-7B, perform adequately in small and medium scenarios but fail at large scale due to context window limitations.}
\textcolor{black}{Taken together, these findings indicate that qualitative performance declines with increasing region size for all models, with GPT-5.2 exhibiting the best scalability and hallucination control, GPT-4.1-mini excelling in moderate contexts, and DeepSeek-V3-685B and Llama3.1-70B emerge as the excellent open-weight alternatives, although worse than OpenAI’s commercial models.}

\subsubsection{\textcolor{black}{Ablation test and comparison with other RAG solutions}}

\textcolor{black}{To validate the proposed architecture, a comprehensive comparison was conducted against other representative RAG solutions. An ablation test was performed to analyze the individual impact of each component on the overall system performance. To this end, the following configurations were evaluated:
}

\begin{enumerate}
    \item \textcolor{black}{Semantic RAG. This configuration corresponds to a naïve RAG approach in which each document is modeled as an NGSI-LD entity. Since this scenario did not include geospatial filtering, the retrieval process was always performed over the full set of available entities, regardless of the region size considered (Small, Medium, or Large). Text-embedding-3-large model with 1024 dimensions was used to generate embeddings and a top-k value of 25 was selected for the retrieval.}

    \item \textcolor{black}{Semantic RAG with reranking. This configuration extended the previous scenario by adding an additional reranking stage based on the ms-marco-MiniLM-L-6-v2 model. In this phase, a top-n of 10 documents was selected, refining the results obtained from the initial retrieval.}
    
    \item \textcolor{black}{Spatial RAG. In this case, only geospatial-based filtering was applied, without incorporating semantic retrieval or reranking. For each scenario, the model received the complete set of entities from the region.}
    
    \item \textcolor{black}{Spatial RAG combined with Semantic RAG. This architecture extended the Spatial RAG scenario by incorporating a semantic retrieval stage after geospatial filtering.}
    
    \item \textcolor{black}{Spatial RAG combined with Semantic RAG and reranking. Finally, this configuration added a reranking stage to the previous scenario.}
\end{enumerate}

\begin{figure}
    \centering
    \includegraphics[width=0.8\linewidth]{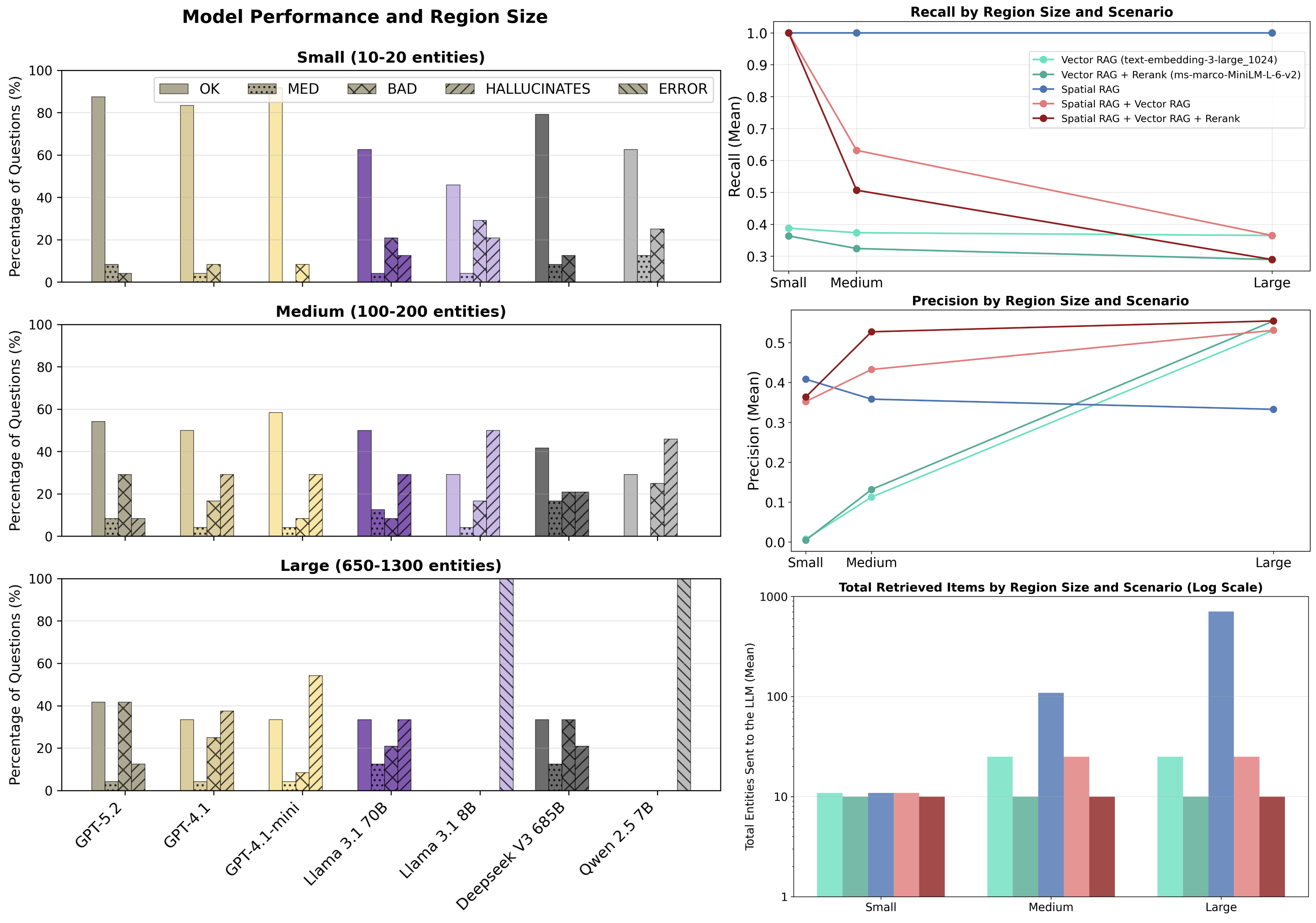}
    \vspace{-5pt}
    \caption{\textcolor{black}{Models' performance in different region sizes (left). Precision, Recall, and Number of entities retrieved in different RAG solutions (right). Recall in Spatial RAG is one because it retrieves all the entities from the region.}}
    \vspace{-5pt}
    \label{fig:recall_precision}
\end{figure}

\textcolor{black}{Figure \ref{fig:recall_precision} presents the results after executing the 24 questions in the five configurations. We obtained precision, recall, and the total number of retrieved documents. Small regions are where the impact of incorporating geospatial filtering is most clearly observed. Without this mechanism (in the Semantic RAG and Semantic RAG with reranking scenarios [1–2]), the system fails to retrieve the relevant entities. In these small regions the effect of adding Semantic RAG, with or without reranking, on top of Spatial RAG (scenarios 3-4) is marginal, since the LLM can directly process all available documents, as observed in previous experiments. Across all scenarios, the same number of documents is sent to the LLM, but the low precision and recall make Semantic RAG, with or without reranking (1-2), unviable in this context. Medium-sized regions are where the most pronounced differences between configurations emerge. A similar behavior to that described above is observed regarding the inclusion or exclusion of Spatial RAG. However, using Spatial RAG alone (3) begins to significantly increase the number of retrieved entities passed to the LLM, leading to scenarios in which the model is no longer able to process them adequately. At this point, the inclusion of Semantic RAG, with or without reranking (4-5), complements Spatial RAG and provides a balance between the number of entities that the LLM can handle and the resulting precision and recall. For large regions, the results are identical with or without Spatial RAG, since geospatial filtering has no effect and all entities stored in the context broker are retrieved (1-2, 4-5). In a pure Spatial RAG scenario, precision and recall reach sufficiently high values, but the number of retrieved entities is also large, already exceeding the operational limits of an LLM.}

\textcolor{black}{Overall results show that the inclusion of Spatial RAG (3) improves precision and recall, while the addition of Semantic RAG (4), potentially combined with reranking (5), enables the retrieval of a number of documents that LLMs can process efficiently. However, it is crucial to carefully select the region size used for geospatial filtering, as large regions render Semantic RAG, even with reranking, ineffective.}

\subsubsection{Measuring latency on different contextual information}
\label{ref_results_latency} 

\begin{figure}
    \centering
    \includegraphics[width=0.9\linewidth]{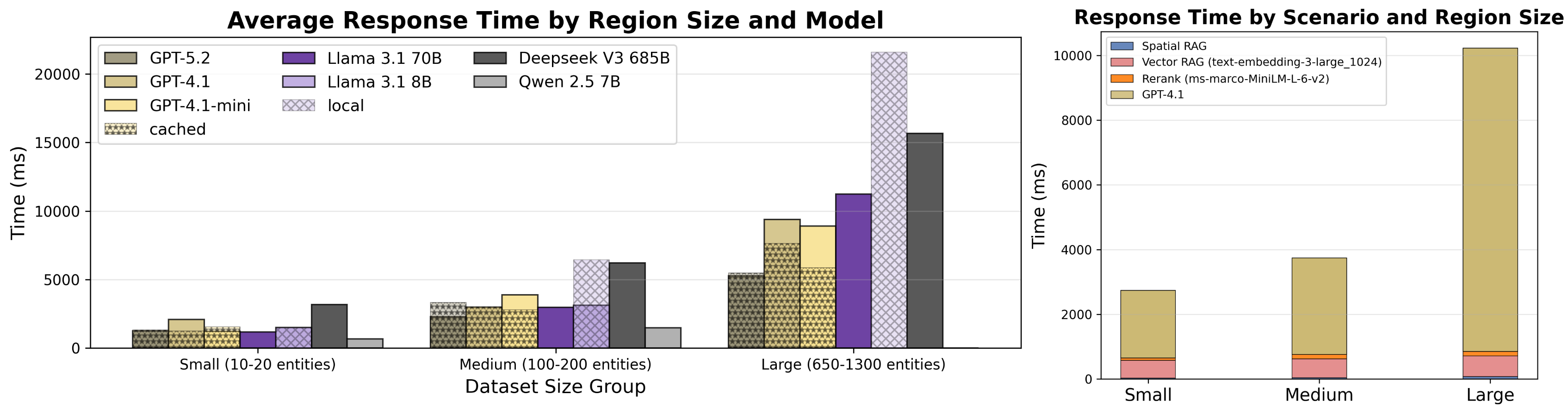}
    \vspace{-5pt}
    \caption{\textcolor{black}{Response times of different models (left) and different RAG solutions (right) in the three region sizes. Llama-3.1-8B and Qwen-2.5-7B were unable to execute large-scale experiments because of insufficient available memory in the context window.}}
    \vspace{-5pt}
    \label{fig:times}
\end{figure}

\textcolor{black}{In this analysis we evaluated the response time of LLMs and of the different components of the proposed RAG architecture (Figure \ref{fig:times}). The models were executed on OpenAI servers (GPT-5.2, GPT-4.1, GPT-4.1-mini), on Hugging Face inference providers (Novita, Together AI and Hyperbolic for models Llama3.1-8B and DeepseekV3-685B, Qwen2.5-7B and Llama3.1-70B, respectively), and on a local server with an A100 Nvidia GPU of 40 GB VRam for Llama3.1-8B. A linear growth in response time is observed as the region size increases across all models, indicating that the amount of context affects not only response quality but also delay. All models exhibit a minimum delay of between one and two seconds in the small region, where they only handle between 10 and 20 entities, which indicates that this will likely be the minimum operational latency. As for the models hosted by OpenAI, they behave differently than anticipated, since GPT-4.1 and GPT-4.1-mini exhibit longer response times than GPT-5.2 in all three scenarios, especially in large regions where the difference becomes more pronounced. One possible reason is that OpenAI may be using higher-performing GPUs for GPT-5.2 or that GPT-4 executions occur less frequently due to batching. It is noteworthy that GPT-4.1-mini, despite its efficiency-oriented design, is not always the fastest and is quite close to GPT-4.1, and even slower than GPT-4.1 in the intermediate scenario. This suggests that reducing model capacity does not necessarily translate into a proportional improvement in latency under more demanding workloads in commercial providers. In the case of open models, greater dispersion in response times can be observed. Llama 3.1 8B and Qwen 2.5 7B maintain relatively low latencies in small and medium regions, comparable to or even lower than some proprietary models, which positions them as efficient alternatives for smaller-scale scenarios. However, as the region size increases, Llama 3.1 70B shows a notable increase in response time, reflecting the direct impact of model size on inference latency. In addition, the smaller models (Llama 3.1 8B and Qwen 2.5 7B) were not able to run the large scenario due to context window limitations. The most extreme case is DeepSeek V3 685B, which exhibits the highest response times across all region sizes, with a very pronounced growth in the large scenario. This result highlights that, although very large models can provide advantages in quality or reasoning capability, their integration into RAG architectures must be carefully assessed when latency is a critical factor.}

\textcolor{black}{We evaluated the impact of prompt caching on OpenAI models. The models can cache the common and static components at the beginning of the prompt, in our case the system prompt and NGSI-LD entities. For GPT 5.2, no improvements were observed, with identical results across scenarios and even slightly worse performance in the intermediate case. In contrast, GPT 4.1 and GPT 4.1 mini exhibit a noticeable improvement, particularly in the large scenario, where a greater volume of raw information can be cached, leading to response time reductions of up to 30\% in GPT 4.1 mini. Nevertheless, LLM response times remain high in this scenario. We also run Llama 3.1-8B locally on an Nvidia A100 GPU with 40GB of VRAM. No significant improvement in overall response time is observed, since the main source of delay comes from the model’s own inference process and not from the network. However, using local models does help prevent vendor lock-in and can be particularly suitable for environments without internet connection or where data must remain within the provider’s own infrastructure for security or compliance reasons}

\textcolor{black}{Another key dimension in the evaluation is execution cost, since there is a direct relationship between model size and price per token. In this respect, the differences are very significant. For example, Llama 3.1 70B has an approximate cost eight times higher than Llama 3.1 8B, and it can be up to about 350 times more expensive when compared with GPT-5.2 in terms of output tokens.}\footnote{\textcolor{black}{As of April 2026, prices per million tokens are: GPT-5.2 1.75\$ input, 14.00\$ output; GPT-4.1 2.00\$ and 8.00\$; GPT-4.1-mini 0.40\$ and 1.60\$ (data from OpenAI); Llama 3.1 8B 0.02\$ and 0.05\$ output; Llama 3.1 70B 0.40\$ and 0.40\$; Qwen 2.5 7B 0.04\$ and 0.10\$; DeepSeek V3 685B 1.25\$ and 1.25\$ (data from OpenRouter)}} \textcolor{black}{This disparity makes cost a determining variable when designing and deploying a RAG architecture in a realistic environment at scale.These results reinforce the need to jointly consider latency, cost, and model capacity. While proprietary models offer optimized performance and more controlled latency, open-weight models stand out for their economic efficiency, making them attractive candidates for large-scale deployments or scenarios where budget is a significant constraint.}

\textcolor{black}{Figure \ref{fig:times} shows the aggregated response time for the different scenarios and region sizes, broken down by the main system components. It can be clearly observed that the overall execution time is dominated by the LLM latency, which constitutes the primary bottleneck and is also the only component whose cost increases substantially when moving from small to large scenarios. Semantic RAG represents the second most relevant contribution to the total response time, yet it remains largely stable across scenarios. Both Semantic RAG and the reranking show small increases of time when the scenario grows, indicating scalable behavior that is largely independent of the region size. Similarly, the CB component does not introduce any noticeable temporal overhead in any of the evaluated scenarios.}

\textcolor{black}{The impact of Spatial RAG, one of the key novelties of this study, is particularly noteworthy while its contribution to the overall response time is practically negligible across all scenarios and does not scale with region size. This demonstrates that incorporating spatial information can be achieved without compromising system performance, reinforcing its practical feasibility and its value as a methodological contribution. In addition, the proposed systems are inherently scalable. FIWARE provides Context Broker federation mechanisms, which makes the approach suitable for high volume of data scenarios. The key aspect is keeping the effective context bounded. In urban settings, geospatial filtering naturally limits the total number of relevant entities, while the subscription mechanism ensures that platforms only receive updates for entities that change. This avoids the need to periodically pull the entire database, significantly reducing communication overhead and improving overall system efficiency.}

\subsubsection{Effect of questions on the LLMs response time}

\textcolor{black}{Since the main limitation in terms of latency is the LLM, we conducted a few additional experiments to analyze how the number of entities and the type of question affect it. To do so,} we focused on assessing the latency associated with retrieving entities from the Orion-LD context broker and subsequently obtaining responses from OpenAI GPT-4.1 API. 

The logged latency values included both the time required to fetch the relevant entities from the context broker and the time associated with querying the LLM. After each query was repeated 10 times and the latencies recorded, the maximum entities limit parameter was increased from 10 to 100, and eventually to 650 PoIs.

We deliberately sought to cover a broad spectrum of \textcolor{black}{new} prompts, including general, open-ended questions such as ``What can I visit in Madrid?'' as well as more narrowly constrained prompts such as ``What can I visit today in Madrid that costs between 10 and 20€?''. Furthermore, we incorporated queries referencing well-known entities where relevant PoI data was expected to be present such as ``Do you know if I can visit Museo del Prado?'', as well as deliberately misleading or contextually unsuitable inquiries ``Do you know if I can visit the Eiffel Tower in Madrid?'' to test scenarios where the answer was unequivocally unavailable or incorrect. By constructing a diverse query set along these dimensions, ranging from generic to highly specific, and from presumably correct to manifestly invalid, we aimed to assess how the complexity, specificity, and foreknowledge of a query’s validity might influence the UFM’s overall latency performance. The queries used in this phase of the experiment are labeled as latency questions (Q) and are included in Table \ref{tab:table-ql}.

\begin{table}[]
\centering
\tiny
\caption{Questions to evaluate the latency of the real-time spatial RAG system}
\vspace{-10pt}
\label{tab:table-ql}
\begin{tabular}{l|l}
\hline
Q\# & Question \\ \hline
Q1 & What can I visit today in Madrid? \\
Q2 & I want to recommend to tourists some places that have a high number of attendance? \\
Q3 & What can I visit today in Madrid that costs between 10 and 20€? \\
Q4 & \begin{tabular}[c]{@{}l@{}}What can I visit today in Madrid that is free of charge and has a ratio of occupancy of \\ less than 10\%?\end{tabular} \\
Q5 & \begin{tabular}[c]{@{}l@{}}What can I visit today in Madrid that is free of charge and has a ratio of occupancy of \\ less than 10\%? You will know the occupancy percentage with the ratio between the \\ occupancy over the capacity of the PoI.\end{tabular} \\
Q6 & Do you know if I can visit Museo del Prado? \\
Q7 & Do you know if I can visit the Eiffel Tower in Madrid? \\ \hline
\end{tabular}
\end{table}

\begin{figure}
    \centering
    \includegraphics[width=0.9\linewidth]{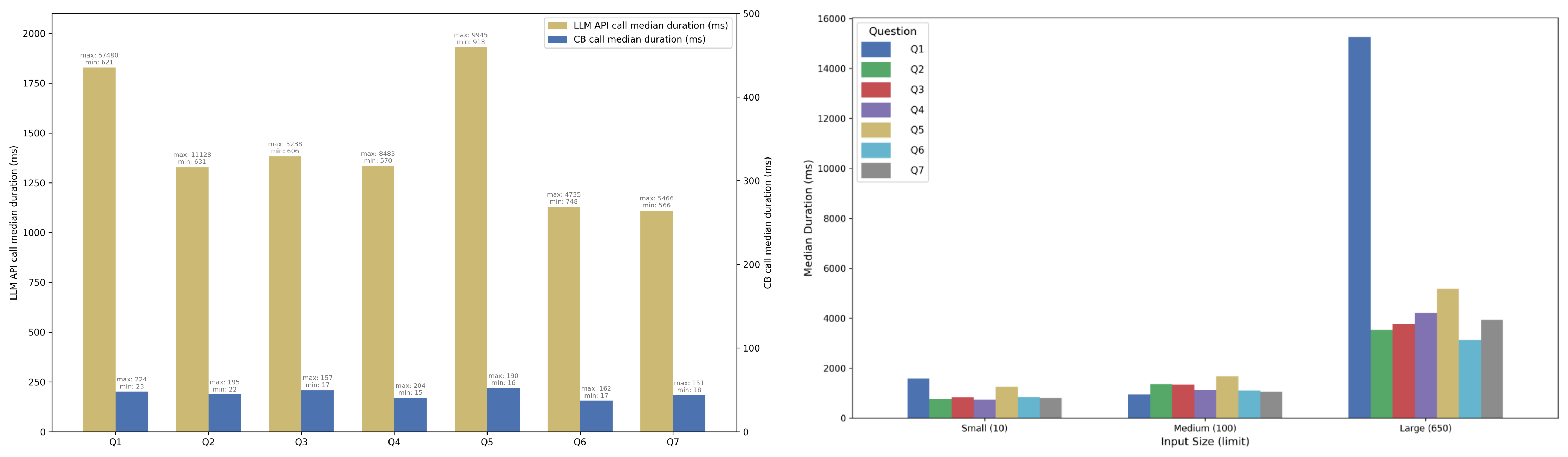}
    \vspace{-10pt}
    \caption{GPT-4.1 and context broker API median times (ms) results for prompts (left); and disaggregated median times of GPT-4.1 + context broker call times (ms) for the three region sizes (right). }
    \label{fig:time-specific-questions}
    \vspace{-10pt}
\end{figure}

For our analysis in these experiments, we chose to report the median of the 10 runs per question rather than mean latency because our response‐time distributions are highly skewed and median is more robust to occasional network or processing outliers. Figure \ref{fig:time-specific-questions}  shows a dual‐axis bar chart with gold bars for the LLM API call median durations plotted against the left y-axis, scaled from $1000$ to $2100$ ms, and blue bars for the context broker median durations against the right y-axis, scaled from $0$ to $500$ ms. Superimposed on every bar are its observed minimum and maximum values. This annotation scheme reveals that, although the LLM medians vary from about $1109-1128$ ms on the simplest lookups (Q6/Q7 respectively) to roughly $1928$ ms for the most complex constraint (Q5), the absolute worst‐case latencies can be much higher (for example, a $57480$ ms maximum on Q1), while minimums remain well below the median ($566$–$918$ ms). This variation in the measurements is mainly due to fluctuations in the load on OpenAI's servers, where multiple users are accessing concurrently.

These results reveal that for every prompt (Q1-Q7) the LLM’s latency is more than an order of magnitude larger than the context broker’s times even when comparing the minimum delay in the LLMs. It is clear that LLM-driven round trips not only exhibit substantially higher typical latencies than context broker queries but also suffer from far greater variability and pronounced long-tail delays, whereas the context broker remains highly predictable and low-variance.

Queries Q1 and Q5 exhibit the highest LLM median latencies because each imposes substantial, albeit different, processing burdens on the model. In the case of Q1, the wholly open‐ended nature of the prompt forces the LLM to generate longer answers, impacting in the number of tokens generated and consequently on the delay. By contrast, Q5 is tightly constrained, multi-step requirement—identifying free points of interest whose occupancy ratio falls below ten percent—compels the LLM to parse and apply a quantitative ratio condition, effectively simulating an on-the-fly calculation in natural language. Together, these results underscore how both broad exploratory queries and complex quantitative constraints can substantially inflate end-to-end LLM response times.

In contrast, queries Q6 and Q7 yield markedly lower median latencies because they reduce the problem to a simple fact‐check rather than open‐ended retrieval or multi‐step reasoning. Whether the model confirms the existence of the Museo del Prado (Q6) or correctly rejects the possibility of an ``Eiffel Tower in Madrid'' (Q7), it needs only to retrieve or verify a single entity and generate a brief affirmation or negation, minimizing both candidate filtering and token generation overhead. As a result, these single‐lookup prompts consistently complete in roughly a fraction of the time required by more complex or broadly scoped queries.

Figure \ref{fig:time-specific-questions} disaggregates the combined median latencies of the LLM API plus context broker calls for each of our seven prompts, plotted at three PoI limits: small (10), medium (100), and large (650). Although the context broker’s share is almost imperceptible in addition to the bulk processing time of the LLM, we include every query at each scale to show how feeding ever-larger context payloads of the model drives end-to-end latency. Across all three limits, the fully open-ended Q1 (``What can I visit today in Madrid?'') is the slowest: broad requests force the model to scan and rank many candidate locations, and that cost grows steeply as you expand from 100 to 650 PoIs. Moreover, as will be seen in the correctness test, when many inputs that meet the conditions are passed to the LLM, it tends to hallucinate and generate very long responses, repeatedly mentioning the same entities.


\textcolor{black}{Results reveal that the current latency of LLMs makes their use in hard real-time settings impractical at this stage. Therefore, further research on LLM compression, improving the capabilities of smaller models, and their deployment at the edge is required to move toward hard real-time and actuator-in-the-loop scenarios, including safety, fault tolerance, and control guarantees.}

\section{Conclusions and future work}

The growing success of Foundation Models in industry, along with their potential for application across an increasing number of domains, highlights the need to standardize their integration into urban environments, taking into account both the components involved and the data flow. In this article, we propose a reference architecture for implementing RAGs in urban environments, as well as an implementation based on FIWARE technology as a comprehensive solution for deploying Foundation Model-based systems in smart cities. This architecture meets all the fundamental requirements of urban environments, including connectivity with IoT devices and third-party systems, scalability, security, and real-time updates, all enabled through geospatial and real-time searches with relationships between entities.

FIWARE, already widely validated in the literature as a robust platform for developing smart city solutions, is presented here as a technology enabler for the inclusion of RAG-based systems in urban contexts. However, although Foundation Models are rapidly evolving, with improvements in speed and larger context windows, significant limitations still remain.

Experiments conducted in three scenarios (with 10\textcolor{black}{--20}, 100\textcolor{black}{--200}, and 600\textcolor{black}{--1300} entities loaded into the model) show that the main bottleneck lies in the LLM itself, which experiences increased generation times when processing large volumes of information. In addition, a higher tendency to error has been observed as the amount of processed city data increases, even when these data fit within the model context window. However, the results obtained with a moderate number of entities are promising, demonstrating that it is currently feasible to implement RAG systems in urban environments, provided that a careful and efficient selection of relevant information is carried out. \textcolor{black}{Future work should address the scalability limitations of the current evaluation, which does not yet reach volumes of millions of entities, although our proposal is designed to mitigate this issue through horizontal scalability of the components used, the spatial decoupling and the filtering mechanisms that operate at the context broker level, ensuring that only geographically and semantically relevant entities are forwarded to the RAG pipeline.}

\textcolor{black}{The architecture presented in this study is use case agnostic, as all the FIWARE components involved are inherently generic \cite{9963753}. This is shown through two completely different use cases: tourism, and traffic and street lighting management. However,} as future lines of work, we propose validating the architecture across a wider variety of scenarios, \textcolor{black}{and urban topologies. Moreover we plan to include the measurement of additional aspects of human perception in the evaluation, such as answer faithfulness, user satisfaction, information density or helpfulness with inter-annotator agreement to validate qualitative insights.} Additionally, the potential of providing the LLM with the ontology associated with each type of entity, as defined in the Smart Data Models, has not yet been explored. This structured and natural language information could help the model more accurately interpret the meaning of each property and the relationships between entities \textcolor{black}{and would enable the resolution of multi-hop queries through the implementation of an MCP service}. Another promising line of research involves investigating the use of LLMs for the automatic generation of queries in NGSI-LD format, or alternatively, training specialized (fine-tuned) models capable of translating natural language queries into that format more efficiently and accurately. Furthermore, the LLM’s ability to autonomously navigate through the entity graph has not been thoroughly evaluated. This functionality could enable dynamic context expansion, \textcolor{black}{multi-hop query resolution and} potentially improving system performance in complex reasoning, information retrieval, and task execution scenario adding agentic capabilities to the system.


Regarding accuracy and scalability, a possible improvement path would be to implement a map-reduce strategy \textcolor{black}{\cite{zhang2025recursive}}. In cases of overly general queries, the search could be broken down into subqueries by spatial areas, performing an independent request for each of them, and finally composing the global response from the partial results. This strategy could facilitate scalability in terms of the number of manageable entities and improve the quality of the generated responses.

\begin{acks}

This work was supported by the Spanish Agencia Estatal de Investigación under Grant FUN4DATE (PID2022-136684OB-C22), by TUCAN6-CM (TEC-2024/COM460) funded by CM (ORDEN 5696/2024) and by the BRIDGE-AI (Grant 101299050) project funded by the European Union's Horizon Europe research and innovation program.

\end{acks}


\bibliographystyle{ACM-Reference-Format}

\bibliography{biblio}

@misc{naveed2024comprehensiveoverviewlargelanguage,
      title={A Comprehensive Overview of Large Language Models}, 
      author={Humza Naveed and Asad Ullah Khan and Shi Qiu and Muhammad Saqib and Saeed Anwar and Muhammad Usman and Naveed Akhtar and Nick Barnes and Ajmal Mian},
      year={2024},
      eprint={2307.06435},
      archivePrefix={arXiv},
      primaryClass={cs.CL},
}

@inproceedings{ovadia-etal-2024-fine,
    title = "Fine-Tuning or Retrieval? Comparing Knowledge Injection in {LLM}s",
    author = "Ovadia, Oded  and
      Brief, Menachem  and
      Mishaeli, Moshik  and
      Elisha, Oren",
    editor = "Al-Onaizan, Yaser  and
      Bansal, Mohit  and
      Chen, Yun-Nung",
    booktitle = "Proceedings of the 2024 Conference on Empirical Methods in Natural Language Processing",
    month = nov,
    year = "2024",
    doi = "10.18653/v1/2024.emnlp-main.15",
    pages = "237--250",
}

@inproceedings{10.5555/3495724.3496517,
author = {Lewis, Patrick and Perez, Ethan and Piktus, Aleksandra and Petroni, Fabio and Karpukhin, Vladimir and Goyal, Naman and K\"{u}ttler, Heinrich and Lewis, Mike and Yih, Wen-tau and Rockt\"{a}schel, Tim and Riedel, Sebastian and Kiela, Douwe},
title = {Retrieval-augmented generation for knowledge-intensive NLP tasks},
year = {2020},
isbn = {9781713829546},
publisher = {Curran Associates Inc.},
address = {Red Hook, NY, USA},
booktitle = {Proceedings of the 34th International Conference on Neural Information Processing Systems},
articleno = {793},
numpages = {16},
location = {Vancouver, BC, Canada},
series = {NIPS '20}
}

@inproceedings{10.1145/3637528.3671470,
author = {Fan, Wenqi and Ding, Yujuan and Ning, Liangbo and Wang, Shijie and Li, Hengyun and Yin, Dawei and Chua, Tat-Seng and Li, Qing},
title = {A Survey on RAG Meeting LLMs: Towards Retrieval-Augmented Large Language Models},
year = {2024},
isbn = {9798400704901},
publisher = {Association for Computing Machinery},
address = {New York, NY, USA},
doi = {10.1145/3637528.3671470},
booktitle = {Proceedings of the 30th ACM SIGKDD Conference on Knowledge Discovery and Data Mining},
pages = {6491–6501},
numpages = {11},
location = {Barcelona, Spain},
series = {KDD '24}
}

@ARTICLE{9086495,
  author={Kirimtat, Ayca and Krejcar, Ondrej and Kertesz, Attila and Tasgetiren, M. Fatih},
  journal={IEEE Access}, 
  title={Future Trends and Current State of Smart City Concepts: A Survey}, 
  year={2020},
  volume={8},
  number={},
  pages={86448-86467},
  doi={10.1109/ACCESS.2020.2992441}}

@article{WEIL2023104862,
title = {Urban Digital Twin Challenges: A Systematic Review and Perspectives for Sustainable Smart Cities},
journal = {Sustainable Cities and Society},
volume = {99},
pages = {104862},
year = {2023},
issn = {2210-6707},
doi = {https://doi.org/10.1016/j.scs.2023.104862},
author = {Charlotte Weil and Simon Elias Bibri and Régis Longchamp and François Golay and Alexandre Alahi},
}

@misc{zhang2025urbangeneralintelligencereview,
      title={Towards Urban General Intelligence: A Review and Outlook of Urban Foundation Models}, 
      author={Weijia Zhang and Jindong Han and Zhao Xu and Hang Ni and Tengfei Lyu and Hao Liu and Hui Xiong},
      year={2025},
      eprint={2402.01749},
      archivePrefix={arXiv},
      primaryClass={cs.CY},
}

@article{exsy.12753,
author = {Haque, A. K. M. Bahalul and Bhushan, Bharat and Dhiman, Gaurav},
title = {Conceptualizing smart city applications: Requirements, architecture, security issues, and emerging trends},
journal = {Expert Systems},
volume = {39},
number = {5},
pages = {e12753},
doi = {https://doi.org/10.1111/exsy.12753},
eprint = {https://onlinelibrary.wiley.com/doi/pdf/10.1111/exsy.12753},
year = {2022}
}

@article{JAVED2022103794,
title = {Future smart cities: requirements, emerging technologies, applications, challenges, and future aspects},
journal = {Cities},
volume = {129},
pages = {103794},
year = {2022},
issn = {0264-2751},
doi = {https://doi.org/10.1016/j.cities.2022.103794},
author = {Abdul Rehman Javed and Faisal Shahzad and Saif ur Rehman and Yousaf Bin Zikria and Imran Razzak and Zunera Jalil and Guandong Xu},
}

@inproceedings{10.1145/3637528.3671453,
author = {Zhang, Weijia and Han, Jindong and Xu, Zhao and Ni, Hang and Liu, Hao and Xiong, Hui},
title = {Urban Foundation Models: A Survey},
year = {2024},
isbn = {9798400704901},
publisher = {Association for Computing Machinery},
address = {New York, NY, USA},
doi = {10.1145/3637528.3671453},
booktitle = {Proceedings of the 30th ACM SIGKDD Conference on Knowledge Discovery and Data Mining},
pages = {6633–6643},
numpages = {11},
location = {Barcelona, Spain},
series = {KDD '24}
}

@article{auto-2019-0039,
title = {An architecture of an Intelligent Digital Twin in a Cyber-Physical Production System},
title = {},
author = {Behrang Ashtari Talkhestani and Tobias Jung and Benjamin Lindemann and Nada Sahlab and Nasser Jazdi and Wolfgang Schloegl and Michael Weyrich},
pages = {762--782},
volume = {67},
number = {9},
journal = {at - Automatisierungstechnik},
doi = {doi:10.1515/auto-2019-0039},
year = {2019},
lastchecked = {2025-04-27}
}

@inproceedings{NEURIPS2024_1403ab1a,
 author = {Dong, Zican and Li, Junyi and Men, Xin and Zhao, Wayne Xin and Wang, Bingning and Tian, Zhen and Chen, Weipeng and Wen, Ji-Rong},
 booktitle = {Advances in Neural Information Processing Systems},
 editor = {A. Globerson and L. Mackey and D. Belgrave and A. Fan and U. Paquet and J. Tomczak and C. Zhang},
 pages = {10320--10347},
 publisher = {Curran Associates, Inc.},
 title = {Exploring Context Window of Large Language Models via Decomposed Positional Vectors},
 volume = {37},
 year = {2024}
}

@inbook{doi:10.1049/PBBE005E_ch10,
author = {Martin Alvarez-Espinar  and Iris Yuping Ren  and Suhail Khan },
title = {Smart governance and e-public administration in smart cities},
booktitle = {Smart Cities for Inclusive Innovation},
chapter = {Chapter 10},
pages = {199-244},
doi = {10.1049/PBBE005E_ch10},
year = {2024}
}

@article{ARSLAN20243781,
title = {A Survey on RAG with LLMs},
journal = {Procedia Computer Science},
volume = {246},
pages = {3781-3790},
year = {2024},
note = {28th International Conference on Knowledge Based and Intelligent information and Engineering Systems (KES 2024)},
issn = {1877-0509},
doi = {https://doi.org/10.1016/j.procs.2024.09.178},
author = {Muhammad Arslan and Hussam Ghanem and Saba Munawar and Christophe Cruz},
}

@INPROCEEDINGS{10569238,
  author={Perković, Gabrijela and Drobnjak, Antun and Botički, Ivica},
  booktitle={2024 47th MIPRO ICT and Electronics Convention (MIPRO)}, 
  title={Hallucinations in LLMs: Understanding and Addressing Challenges}, 
  year={2024},
  volume={},
  number={},
  pages={2084-2088},
  doi={10.1109/MIPRO60963.2024.10569238}}

@ARTICLE{10654534,
  author={Xiong, Haoyi and Bian, Jiang and Li, Yuchen and Li, Xuhong and Du, Mengnan and Wang, Shuaiqiang and Yin, Dawei and Helal, Sumi},
  journal={IEEE Transactions on Services Computing}, 
  title={When Search Engine Services Meet Large Language Models: Visions and Challenges}, 
  year={2024},
  volume={17},
  number={6},
  pages={4558-4577},
  doi={10.1109/TSC.2024.3451185}}

@misc{zhao2024retrievalaugmentedgenerationaigeneratedcontent,
      title={Retrieval-Augmented Generation for AI-Generated Content: A Survey}, 
      author={Penghao Zhao and Hailin Zhang and Qinhan Yu and Zhengren Wang and Yunteng Geng and Fangcheng Fu and Ling Yang and Wentao Zhang and Jie Jiang and Bin Cui},
      year={2024},
      eprint={2402.19473},
      archivePrefix={arXiv},
      primaryClass={cs.CV},
}

@misc{nogueira2020passagererankingbert,
      title={Passage Re-ranking with BERT}, 
      author={Rodrigo Nogueira and Kyunghyun Cho},
      year={2020},
      eprint={1901.04085},
      archivePrefix={arXiv},
      primaryClass={cs.IR},
}

@inproceedings{10.1145/3397271.3401075,
author = {Khattab, Omar and Zaharia, Matei},
title = {ColBERT: Efficient and Effective Passage Search via Contextualized Late Interaction over BERT},
year = {2020},
isbn = {9781450380164},
publisher = {Association for Computing Machinery},
address = {New York, NY, USA},
doi = {10.1145/3397271.3401075},
booktitle = {Proceedings of the 43rd International ACM SIGIR Conference on Research and Development in Information Retrieval},
pages = {39–48},
numpages = {10},
location = {Virtual Event, China},
series = {SIGIR '20}
}

@misc{fiware2023fiware4cities,
  author = {{FIWARE Foundation}},
  title = {{FIWARE 4 CITIES}},
  year = {2023},
  publisher = {FIWARE Foundation e. V.},
  address = {Berlin},
}

@article{carvajal2024enhancing,
  title={Enhancing industrial digitalisation through an adaptable component for bridging semantic interoperability gaps},
  author={Carvajal-Flores, Diego F and Abril-Jim{\'e}nez, Patricia and Buhid, Eduardo and Fico, Giuseppe and Cabrera Umpi{\'e}rrez, Mar{\'\i}a Fernanda},
  journal={Applied Sciences},
  volume={14},
  number={6},
  pages={2309},
  year={2024},
  publisher={MDPI},
  doi={10.3390/app14062309}
}

@Article{agriculture13051005,
AUTHOR = {Emmi, Luis and Fernández, Roemi and Gonzalez-de-Santos, Pablo and Francia, Matteo and Golfarelli, Matteo and Vitali, Giuliano and Sandmann, Hendrik and Hustedt, Michael and Wollweber, Merve},
TITLE = {Exploiting the Internet Resources for Autonomous Robots in Agriculture},
JOURNAL = {Agriculture},
VOLUME = {13},
YEAR = {2023},
NUMBER = {5},
ARTICLE-NUMBER = {1005},
ISSN = {2077-0472},
DOI = {10.3390/agriculture13051005}
}

@InProceedings{10.1007/978-3-031-60796-7-12,
author="De Benedictis, Alessandra
and Rocco di Torrepadula, Franca
and Somma, Alessandra",
editor="Lotfian, Maryam
and Starace, Luigi Libero Lucio",
title="A Digital Twin Architecture for Intelligent Public Transportation Systems: A FIWARE-Based Solution",
booktitle="Web and Wireless Geographical Information Systems",
year="2024",
publisher="Springer Nature Switzerland",
address="Cham",
pages="165--182",
isbn="978-3-031-60796-7"
}

@article{CONDE2022101723,
title = {Applying digital twins for the management of information in turnaround event operations in commercial airports},
journal = {Advanced Engineering Informatics},
volume = {54},
pages = {101723},
year = {2022},
issn = {1474-0346},
doi = {https://doi.org/10.1016/j.aei.2022.101723},
author = {Javier Conde and Andres Munoz-Arcentales and Mario Romero and Javier Rojo and Joaquín Salvachúa and Gabriel Huecas and Álvaro Alonso},
}

@article{loss2023using,
  title={Using FIWARE and blockchain in smart cities solutions},
  author={Loss, Stefano and Singh, Har Preet and Cacho, N{\'e}lio and Lopes, Frederico},
  journal={Cluster Computing},
  volume={26},
  number={4},
  pages={2115--2128},
  year={2023},
  publisher={Springer},
  doi={https://doi.org/10.1007/s10586-022-03732-x},
}

@ARTICLE{9963753,
  author={Conde, Javier and Munoz-Arcentales, Andres and Choque, Johnny and Huecas, Gabriel and Alonso, Álvaro},
  journal={Computer}, 
  title={Overcoming the Barriers of Using Linked Open Data in Smart City Applications}, 
  year={2022},
  volume={55},
  number={12},
  pages={109-118},
  doi={10.1109/MC.2022.3206144}}

@inproceedings{2939672.2939754,
author = {Grover, Aditya and Leskovec, Jure},
title = {node2vec: Scalable Feature Learning for Networks},
year = {2016},
isbn = {9781450342322},
publisher = {Association for Computing Machinery},
address = {New York, NY, USA},
doi = {10.1145/2939672.2939754},
booktitle = {Proceedings of the 22nd ACM SIGKDD International Conference on Knowledge Discovery and Data Mining},
pages = {855–864},
numpages = {10},
location = {San Francisco, California, USA},
series = {KDD '16}
}

@ARTICLE{9346030,
  author={Conde, Javier and Munoz-Arcentales, Andrés and Alonso, Álvaro and López-Pernas, Sonsoles and Salvachúa, Joaquín},
  journal={IEEE Internet Computing}, 
  title={Modeling Digital Twin Data and Architecture: A Building Guide With FIWARE as Enabling Technology}, 
  year={2022},
  volume={26},
  number={3},
  pages={7-14},
  doi={10.1109/MIC.2021.3056923}}

@article{ARAUJO2019250,
title = {Performance evaluation of FIWARE: A cloud-based IoT platform for smart cities},
journal = {Journal of Parallel and Distributed Computing},
volume = {132},
pages = {250-261},
year = {2019},
issn = {0743-7315},
doi = {https://doi.org/10.1016/j.jpdc.2018.12.010},
author = {Victor Araujo and Karan Mitra and Saguna Saguna and Christer Åhlund},
}

@inproceedings{10.1145/3685651.3686661,
author = {Segou, Olga and Skias, Dimitris S. and Velivassaki, Terpsichori-Helen and Zahariadis, Theodore and Pages, Enric and Ramiro, Rub\'{e}n and Rossini, Rosaria and Karkazis, Panagiotis A. and Muniz, Alejandro and Contreras, Luis and del Rio, Alberto and Serrano, Javier and Jimenez, David and Belesioti, Maria and Chochliouros, Ioannis and Vantolas, Spyridon},
title = {NExt generation Meta Operating systems (NEMO) and Data Space: envisioning the future},
year = {2024},
isbn = {9798400709845},
doi = {10.1145/3685651.3686661},
booktitle = {Proceedings of the 4th Eclipse Security, AI, Architecture and Modelling Conference on Data Space},
pages = {41–49},
numpages = {9},
location = {Mainz, Germany},
}

@conference{iotbds23,
author={Kari Kolehmainen and Marco Pirazzi and Juha{-}Pekka Soininen and Juha Backman},
title={Simulation Based Performance Evaluation of FIWARE IoT Platform for Smart Agriculture},
booktitle={Proceedings of the 8th International Conference on Internet of Things, Big Data and Security - IoTBDS},
year={2023},
pages={73-81},
publisher={SciTePress},
organization={INSTICC},
doi={10.5220/0011918700003482},
isbn={978-989-758-643-9},
issn={2184-4976},
}

@inproceedings{10.5555/3666122.3667957,
author = {Li, Jinyang and Hui, Binyuan and Qu, Ge and Yang, Jiaxi and Li, Binhua and Li, Bowen and Wang, Bailin and Qin, Bowen and Geng, Ruiying and Huo, Nan and Zhou, Xuanhe and Ma, Chenhao and Li, Guoliang and Chang, Kevin C.C. and Huang, Fei and Cheng, Reynold and Li, Yongbin},
title = {Can LLM already serve as a database interface? a big bench for large-scale database grounded text-to-SQLs},
year = {2023},
publisher = {Curran Associates Inc.},
address = {Red Hook, NY, USA},
booktitle = {Proceedings of the 37th International Conference on Neural Information Processing Systems},
articleno = {1835},
numpages = {28},
location = {New Orleans, LA, USA},
series = {NIPS '23}
}

@INPROCEEDINGS{10705186,
  author={Vichev, Sergey and Marchev, Angel},
  booktitle={2024 IEEE 12th International Conference on Intelligent Systems (IS)}, 
  title={RAGSQL: Context Retrieval Evaluation on Augmenting Text-to-SQL Prompts}, 
  year={2024},
  volume={},
  number={},
  pages={1-6},
  doi={10.1109/IS61756.2024.10705186}}

@article{10.1145/3296957.3173191,
author = {Lin, Shih-Chieh and Zhang, Yunqi and Hsu, Chang-Hong and Skach, Matt and Haque, Md E. and Tang, Lingjia and Mars, Jason},
title = {The Architectural Implications of Autonomous Driving: Constraints and Acceleration},
year = {2018},
issue_date = {February 2018},
publisher = {Association for Computing Machinery},
address = {New York, NY, USA},
volume = {53},
number = {2},
issn = {0362-1340},
doi = {10.1145/3296957.3173191},
journal = {SIGPLAN Not.},
month = mar,
pages = {751–766},
numpages = {16}
}

@inproceedings{MLSYS2024_42a452cb,
 author = {Lin, Ji and Tang, Jiaming and Tang, Haotian and Yang, Shang and Chen, Wei-Ming and Wang, Wei-Chen and Xiao, Guangxuan and Dang, Xingyu and Gan, Chuang and Han, Song},
 booktitle = {Proceedings of Machine Learning and Systems},
 pages = {87--100},
 title = {AWQ: Activation-aware Weight Quantization for On-Device LLM Compression and Acceleration},
 volume = {6},
 year = {2024}
}

@INPROCEEDINGS{7474180,
  author={Barbieru, Ciprian and Pop, Florin},
  booktitle={2016 IEEE 30th International Conference on Advanced Information Networking and Applications (AINA)}, 
  title={Soft Real-Time Hadoop Scheduler for Big Data Processing in Smart Cities}, 
  year={2016},
  volume={},
  number={},
  pages={863-870},
  doi={10.1109/AINA.2016.122}}

@misc{zhao2024retrievalaugmentedgenerationrag,
      title={Retrieval Augmented Generation (RAG) and Beyond: A Comprehensive Survey on How to Make your LLMs use External Data More Wisely}, 
      author={Siyun Zhao and Yuqing Yang and Zilong Wang and Zhiyuan He and Luna K. Qiu and Lili Qiu},
      year={2024},
      eprint={2409.14924},
      archivePrefix={arXiv},
      primaryClass={cs.CL},
}

@misc{edge2025localglobalgraphrag,
      title={From Local to Global: A Graph RAG Approach to Query-Focused Summarization}, 
      author={Darren Edge and Ha Trinh and Newman Cheng and Joshua Bradley and Alex Chao and Apurva Mody and Steven Truitt and Dasha Metropolitansky and Robert Osazuwa Ness and Jonathan Larson},
      year={2025},
      eprint={2404.16130},
      archivePrefix={arXiv},
      primaryClass={cs.CL},
}

@ARTICLE{10387715,
  author={Pan, Shirui and Luo, Linhao and Wang, Yufei and Chen, Chen and Wang, Jiapu and Wu, Xindong},
  journal={IEEE Transactions on Knowledge and Data Engineering}, 
  title={Unifying Large Language Models and Knowledge Graphs: A Roadmap}, 
  year={2024},
  volume={36},
  number={7},
  pages={3580-3599},
  doi={10.1109/TKDE.2024.3352100}}

@inproceedings{
tang2024multihoprag,
title={MultiHop-{RAG}: Benchmarking Retrieval-Augmented Generation for Multi-Hop Queries},
author={Yixuan Tang and Yi Yang},
booktitle={First Conference on Language Modeling},
year={2024},
}

@inproceedings{
sarthi2024raptor,
title={{RAPTOR}: Recursive Abstractive Processing for Tree-Organized Retrieval},
author={Parth Sarthi and Salman Abdullah and Aditi Tuli and Shubh Khanna and Anna Goldie and Christopher D Manning},
booktitle={The Twelfth International Conference on Learning Representations},
year={2024},
}

@techreport{ Context_Information_Management,
  author = {{European Telecommunications Standards Institute}},
  title = {{Context Information Management (CIM); NGSI-LD API}},
  type = {{ETSI-GS-CIM-009}}, 
  year = {2020},
}

@article{
DaLiouChen2024OpenTI,
  author    = {Da, L. and Liou, K. and Chen, T. and others},
  title     = {Open-ti: Open Traffic Intelligence with Augmented Language Model},
  journal   = {International Journal of Machine Learning \& Cybernetics},
  volume    = {15},
  pages     = {4761--4786},
  year      = {2024},
  doi       = {10.1007/s13042-024-02190-8},
}

@Article{smartcities7060121,
AUTHOR = {Ieva, Saverio and Loconte, Davide and Loseto, Giuseppe and Ruta, Michele and Scioscia, Floriano and Marche, Davide and Notarnicola, Marianna},
TITLE = {A Retrieval-Augmented Generation Approach for Data-Driven Energy Infrastructure Digital Twins},
JOURNAL = {Smart Cities},
VOLUME = {7},
YEAR = {2024},
NUMBER = {6},
PAGES = {3095--3120},
ISSN = {2624-6511},
DOI = {10.3390/smartcities7060121}
}

@InProceedings{10.1007/978-981-99-7962-2-30,
author="Marvin, Ggaliwango
and Hellen, Nakayiza
and Jjingo, Daudi
and Nakatumba-Nabende, Joyce",
editor="Jacob, I. Jeena
and Piramuthu, Selwyn
and Falkowski-Gilski, Przemyslaw",
title="Prompt Engineering in Large Language Models",
booktitle="Data Intelligence and Cognitive Informatics",
year="2024",
publisher="Springer Nature Singapore",
address="Singapore",
pages="387--402",
isbn="978-981-99-7962-2"
}

@article{abs-2407-01219,
  publtype={informal},
  author={Xiaohua Wang and Zhenghua Wang and Xuan Gao and Feiran Zhang and Yixin Wu and Zhibo Xu and Tianyuan Shi and Zhengyuan Wang and Shizheng Li and Qi Qian and Ruicheng Yin and Changze Lv and Xiaoqing Zheng and Xuanjing Huang},
  title={Searching for Best Practices in Retrieval-Augmented Generation},
  year={2024},
  cdate={1704067200000},
  journal={CoRR},
}

@misc{fu2023humanaicollaborativeurbanscience,
      title={Towards Human-AI Collaborative Urban Science Research Enabled by Pre-trained Large Language Models}, 
      author={Jiayi Fu and Haoying Han and Xing Su and Chao Fan},
      year={2023},
      eprint={2305.11418},
      archivePrefix={arXiv},
      primaryClass={cs.HC},
}

@misc{ji2023evaluatingeffectivenesslargelanguage,
      title={Evaluating the Effectiveness of Large Language Models in Representing Textual Descriptions of Geometry and Spatial Relations}, 
      author={Yuhan Ji and Song Gao},
      year={2023},
      eprint={2307.03678},
      archivePrefix={arXiv},
      primaryClass={cs.CL},
}

@article{10.1145/3631937,
author = {Mei, Lang and Mao, Jiaxin and Hu, Juan and Tan, Naiqiang and Chai, Hua and Wen, Ji-Rong},
title = {Improving First-stage Retrieval of Point-of-interest Search by Pre-training Models},
year = {2023},
issue_date = {May 2024},
publisher = {Association for Computing Machinery},
address = {New York, NY, USA},
volume = {42},
number = {3},
issn = {1046-8188},
doi = {10.1145/3631937},
journal = {ACM Trans. Inf. Syst.},
month = dec,
articleno = {74},
numpages = {27}
}

@article{zhang2025recursive,
  title={Recursive Language Models},
  author={Zhang, Alex L and Kraska, Tim and Khattab, Omar},
  journal={arXiv preprint arXiv:2512.24601},
  year={2025}
}

@inproceedings{feng2025urbanllava,
  title={\textcolor{black}{UrbanLLaVA: A Multi-modal Large Language Model for Urban Intelligence}},
  author={\textcolor{black}{Feng, Jie and Wang, Shengyuan and Liu, Tianhui and Xi, Yanxin and Li, Yong}},
  booktitle={\textcolor{black}{Proceedings of the IEEE/CVF International Conference on Computer Vision}},
  pages={\textcolor{black}{6209--6219}},
  year={\textcolor{black}{2025}}
}

@inproceedings{10.1145/3711896.3736878,
author = {\textcolor{black}{Feng, Jie and Liu, Tianhui and Du, Yuwei and Guo, Siqi and Lin, Yuming and Li, Yong}},
title = {\textcolor{black}{CityGPT: Empowering Urban Spatial Cognition of Large Language Models}},
year = {\textcolor{black}{2025}},
isbn = {\textcolor{black}{9798400714542}},
doi = {10.1145/3711896.3736878},
booktitle = {\textcolor{black}{Proceedings of the 31st ACM SIGKDD Conference on Knowledge Discovery and Data Mining V.2}},
pages = {\textcolor{black}{591–602}},
numpages = {\textcolor{black}{12}},
}

@article{wang2025urban,
  title={\textcolor{black}{Urban-R1: Reinforced MLLMs Mitigate Geospatial Biases for Urban General Intelligence}},
  author={\textcolor{black}{Wang, Qiongyan and Zou, Xingchen and Jiang, Yutian and Wen, Haomin and Wei, Jiaheng and Wen, Qingsong and Liang, Yuxuan}},
  journal={\textcolor{black}{arXiv preprint arXiv:2510.16555}},
  year={\textcolor{black}{2025}}
}

@article{zhu2025boundary,
  title={\textcolor{black}{Boundary Prompting: Elastic Urban Region Representation via Graph-based Spatial Tokenization}},
  author={\textcolor{black}{Zhu, Haojia and Jin, Jiahui and Kan, Dong and Shen, Rouxi and Wang, Ruize and Sun, Xiangguo and Zhang, Jinghui}},
  journal={\textcolor{black}{arXiv preprint arXiv:2503.07991}},
  year={\textcolor{black}{2025}}
}

@inproceedings{10.1145/3711896.3737176,
author = {\textcolor{black}{Jin, Jiahui and Song, Yifan and Kan, Dong and Zhu, Haojia and Sun, Xiangguo and Li, Zhicheng and Sun, Xigang and Zhang, Jinghui}},
title = {\textcolor{black}{Urban Region Pre-training and Prompting: A Graph-based Approach}},
year = {\textcolor{black}{2025}},
isbn = {\textcolor{black}{9798400714542}},
publisher = {\textcolor{black}{Association for Computing Machinery}},
address = {\textcolor{black}{New York, NY, USA}},
doi = {10.1145/3711896.3737176},
booktitle = {\textcolor{black}{Proceedings of the 31st ACM SIGKDD Conference on Knowledge Discovery and Data Mining V.2}},
pages = {\textcolor{black}{1071–1082}},
numpages = {\textcolor{black}{12}},
location = {\textcolor{black}{Toronto ON, Canada}},
series = {\textcolor{black}{KDD '25}}
}

@article{yu2025spatial,
  title={\textcolor{black}{Spatial-rag: Spatial retrieval augmented generation for real-world geospatial reasoning questions}},
  author={\textcolor{black}{Yu, Dazhou and Bao, Riyang and Ning, Ruiyu and Peng, Jinghong and Mai, Gengchen and Zhao, Liang}},
  journal={\textcolor{black}{arXiv preprint arXiv:2502.18470}},
  year={\textcolor{black}{2025}}
}

\end{document}